\documentclass{article}

\usepackage[preprint]{neurips_2026}

\usepackage[utf8]{inputenc} 
\usepackage[T1]{fontenc} 
\usepackage{hyperref} 
\usepackage{url} 
\usepackage{booktabs} 
\usepackage{amsfonts} 
\usepackage{nicefrac} 
\usepackage{microtype} 
\usepackage{xcolor} 

\usepackage{amsmath}
\usepackage{amssymb}
\usepackage{mathtools}
\usepackage{amsthm}
\usepackage{multirow}
\usepackage{colortbl}
\usepackage{xcolor}
\usepackage[capitalize, noabbrev]{cleveref}
\usepackage{microtype}
\usepackage{graphicx}
\usepackage{subcaption}
\usepackage{booktabs} 
\usepackage{CJKutf8}
\usepackage{pifont}
\usepackage{wrapfig}
\usepackage{booktabs}
\usepackage[table]{xcolor}
\usepackage{pifont}
\usepackage{graphicx}

\title{Can We Model the Artifacts Explicitly? \\
Disentangle Artifacts via Pairwise Edit Relations for Image Manipulation Localization}

\author{%
  \textbf{Xuekang Zhu}\textsuperscript{1,2},
  \textbf{Kaiwen Feng}\textsuperscript{1},
  \textbf{Ruifeng Wang}\textsuperscript{1},
  \textbf{Xiwen Wang}\textsuperscript{1},
  \textbf{Xiaochen Ma}\textsuperscript{3}
  \\[2pt]
  \textbf{Bo Du}\textsuperscript{1},
  \textbf{Changjiang Jiang}\textsuperscript{2,4},
  \textbf{Chenfan Qu}\textsuperscript{2,5},
  \textbf{Songyu Ye}\textsuperscript{6}
  \\[2pt]
  \textbf{Xia Du}\textsuperscript{7},
  \textbf{Wentao Feng}\textsuperscript{1},
  \textbf{Jian Liu}\textsuperscript{2},
  \textbf{Ji-Zhe Zhou}\textsuperscript{1\ddag}
  \\[4pt]
  \textsuperscript{1} Sichuan University \quad
  \textsuperscript{2} Ant Group
  \\[1pt]
  \textsuperscript{3} The Hong Kong University of Science and Technology \quad
  \textsuperscript{4} Wuhan University
  \\[1pt]
  \textsuperscript{5} South China University of Technology \quad
  \textsuperscript{6} University of Southern California
  \\[1pt]
  \textsuperscript{7} Xiamen University of Technology
}

\begin{document}
\renewcommand{\thefootnote}{\ddag}
\footnotetext{%
Corresponding author: Ji-Zhe Zhou
(\href{mailto:jzzhou@scu.edu.cn}{jzzhou@scu.edu.cn}).
}
  \maketitle

  \begin{abstract}

Image Manipulation Localization (IML) is commonly formulated as a fully supervised learning task that estimates the optimal manipulation mask $y$ for a given image $x$.
In this work, we first reveal the latent nature of artifacts and thus reinterpret IML as a latent-variable problem,
$P(y| x)=\int P(y|z)\,P(z|x)\,dz$,
where $z$ denotes the artifacts.
Following this interpretation, we pinpoint the cause for the current IML models' insufficiency as their implicit artifacts modeling strategy, highlighting the necessity of modeling $z$ in an explicit manner. 
Without direct labels, feature disentanglement is the most appropriate solution for this explicit modeling. Accordingly, we propose a two-stage learning paradigm with the Pairwise Artifacts Learning (PAL) and Standard Localization (SL) phases to estimate $P(z|x)$ and $P (y|z)$ via edit relations.
To support our edit-relation-based learning, we further curate \textit{EditGroup-45K}, a source-anchored dataset organized into edit groups for pair construction.
Extensive experiments show that our PAL paradigm yields consistent improvements across diverse IML architectures, and empirical analyses further verify that PAL does capture artifacts explicitly through feature disentanglement.Code and dataset are available at \url{https://github.com/venus-guangjian/PAL}
\end{abstract}

  \section{Introduction}
\label{sec:intro}
\textit{``What is essential is invisible to the eye" - \underline{\textbf{The Little Prince}}}
\par 
Image Manipulation Localization (IML) is typically formulated as a supervised learning task that estimates the conditional distribution $P(y|x)$, where $y$ is a binary mask indicating manipulated regions, given an input image $x$. The success of this task hinges on the model's ability to capture artifacts, which are subtle feature inconsistencies introduced by manipulation operations. Although decisive to the output mask $y$, artifacts reside primarily in the non-semantic domain and are thereby largely imperceptible to the human eye, rendering direct annotation infeasible.
Consequently, artifacts act as a latent variable, interpreting IML as a latent-variable learning problem governed by the probabilistic decomposition: $P(y|x) = \int P(y|z)P(z|x)dz$, where $z$ represents the underlying manipulation artifacts.

\par
\begin{figure}[t]
\centering

\begin{minipage}[c]{0.44\linewidth}
\centering
\captionof{table}{Comparison of artifacts modeling paradigms. `Seg.' and `Conv.' denote Segformer and ConvNeXt, respectively.}
\label{tab:paradigm_comparison}

\vspace{2mm}

\resizebox{\linewidth}{!}{
\begin{tabular}{l|c|c}
\toprule
\textbf{Model Name} & \textbf{Backbone} & \textbf{Artifacts Modeling} \\
\midrule
PSCC~\cite{liu2022pscc}  & HRNet & Implicit \\
IML-ViT~\cite{ma2023iml} & ViT & Implicit \\
MVSS-Net~\cite{mvsspp_2022} & ResNet & Implicit + Handcrafted \\
CAT-Net~\cite{CAT-Net2022} & HRNet & Implicit + Handcrafted \\
TruFor~\cite{trufor2023} & Segformer & Implicit + Handcrafted \\
Mesorch~\cite{zhu2025mesorch} & Seg. + Conv. & Implicit + Handcrafted \\
\midrule
\rowcolor{gray!15} \textbf{Ours} & \textbf{Any} & \textbf{Explicit} \\
\bottomrule
\end{tabular}
}
\end{minipage}
\hfill
\begin{minipage}[c]{0.54\linewidth}
\centering
\includegraphics[width=\linewidth]{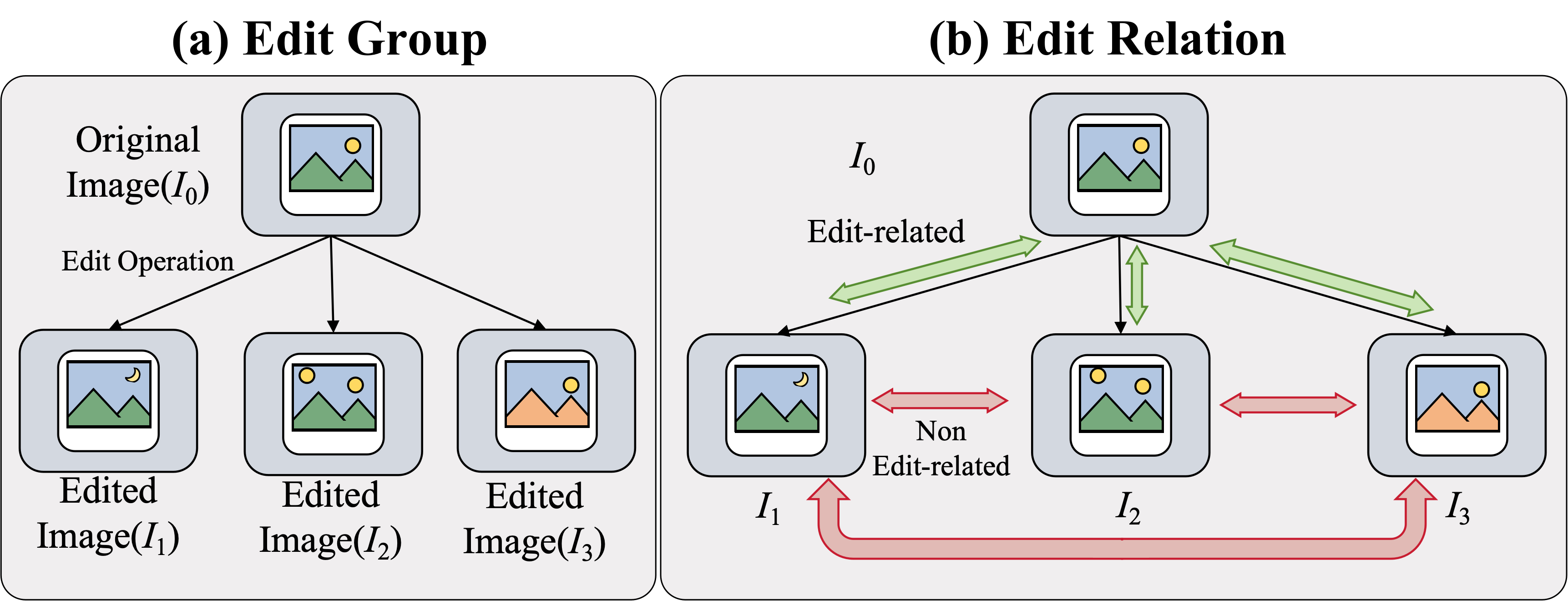}

\vspace{1mm}

\captionof{figure}{\textbf{Edit Groups and Edit Relations.}
(a) Edit Groups are organized as genealogical trees rooted at authentic anchors ($I_0$).
(b) Green and red arrows indicate edit-related and non-edit-related pairs, respectively.}
\label{fig:edit_relations}
\end{minipage}
\vspace{-7mm}
\end{figure}

As Table~\ref{tab:paradigm_comparison} illustrates, to address this latent-variable learning problem, existing IML methods generally adopt a one-stage end-to-end implicit modeling paradigm. Specifically, current solutions attempt to estimate the conditional distribution $P(y|x)$ directly, assuming that the deep model will spontaneously generate representations of the latent artifacts. However, hindered by label noise~\cite{NCL_IML_2023} and semantic shortcuts~\cite{yan2024effort,zhou2025beyond}, such implicit learning methods are highly vulnerable to spurious correlations and are prone to overfitting to semantic biases. To further mitigate these vulnerabilities, many approaches incorporate additional domain expertise via handcrafted feature extractors (e.g., SRM filters~\cite{SRM_2018} or Bayar convolutions~\cite{Bayar_2018}), thereby injecting human priors and constraints into the implicit artifacts learning process. Nevertheless, extensive experiments~\cite{ma2025imdl,du2025forensichubunifiedbenchmark} demonstrate that these rigid and non-adaptive human priors in fact compromise the model's generalization ability across diverse manipulation scenarios.
\par
Considering the above inherent issues of current implicit learning paradigms, a core question for advancing IML arises: \textbf{Can we model the artifacts explicitly?}


The answer to this question lies in disentangling artifacts from the coupled latent space. 
Fundamentally, IML models identify manipulated regions by capturing discrepancies between manipulated and authentic images; however, these discrepancies arise from two components: semantic variations and manipulation artifacts (e.g., non-semantic variations). In other words, while artifacts are the target of interest, they are inherently entangled with semantic features in the latent space. Thus, in the absence of direct annotations for artifacts, disentangling these two components becomes the optimal path for explicit artifacts modeling. Mathematically, \textit{explicit} means that artifacts are introduced as a direct optimization target through relational supervision, rather than being treated as an implicit byproduct of end-to-end mask prediction.
This does not require the stronger assumption that artifacts and semantic factors are statistically independent; instead, it aims to reduce semantic shortcuts and emphasize edit-induced discrepancies.

\par 
To this end, we exploit extensive inter-image relationships and discover that the edit relation between images holds the key to factoring out the artifacts. 
Specifically, as illustrated in Figure~\ref{fig:edit_relations}, the edit relation is the direct correspondence between an authentic image and its manipulated counterpart. Discrepancies in edit-related pairs represent a composite of both artifacts and semantic shifts, whereas discrepancies in non-related pairs originate exclusively from semantic variations.
By contrasting these edit relations, we can effectively suppress the representation of semantic variations, thereby disentangling the latent artifacts $z$. A formal study is also conducted to analyze the disentangled outcomes of this edit relation-based paradigm in Sec.~\ref{app:theory}. Driven by this disentanglement, the intractable end-to-end conditional distribution $p(y|x)$ is decomposed into two manageable sub-processes: explicit artifacts disentanglement $P(z|x)$ and artifact-based localization $P(y|z)$.

Building upon this insight, we therefore propose a two-stage training paradigm consisting of
\textbf{\textit{Pairwise Artifacts Learning} (PAL)} and \textbf{\textit{Standard Localization} (SL)}. The PAL stage is dedicated to explicitly modeling the artifacts posterior $P(z|x)$ by contrasting edit-related and non-related image pairs.
Specifically, we minimize a pairwise binary cross-entropy loss by contrasting edit pairs containing complete artifacts against those with absent artifacts, thereby yielding a PAL-optimized encoder that captures robust, disentangled forensic representations.

Subsequently, the SL stage adapts the PAL-optimized encoder to the pixel-wise manipulation localization objective, i.e., learning the conditional distribution $P(y|z)$ from artifacts to masks. To ensure compatibility and fair comparison with prior IML methods, we adopt the standard training protocol used in the existing IML codebase~\cite{ma2025imdl}. By using the PAL-optimized encoder as backbone initialization, the model starts from an artifact-aware representation that facilitates estimating $P(z|x)$, thereby simplifying the downstream optimization from the entangled end-to-end objective $P(y|
x)$ to the more direct conditional distribution $P(y|z)$ and reducing reliance on semantic shortcuts.

However, a bottleneck for implementing this approach is the scarcity of large-scale datasets containing explicit edit pairs. To establish a training basis for our proposed approach, we curate \textbf{\textit{EditGroup-45K}}, a large-scale dataset specifically organized into edit groups. This dataset comprises 52K authentic images and 232K manipulated images across 45K edit groups, totaling approximately 284K images. Crucially, each group maintains a strict correspondence, consisting of at least one authentic image associated with its specific manipulated counterparts. Unlike traditional settings where samples are treated in isolation, this group-based architecture provides the massive-scale explicit relationships necessary to ground the artifacts modeling. More details are provided in Sec.~\ref{sec:dataset}.

Leveraging the EditGroup-45K dataset, our PAL empowers diverse backbones to explicitly capture generalized artifacts, thereby establishing a superior initialization for the subsequent localization adaptation. Extensive experimental results demonstrate that our paradigm serves as a universal performance booster. It yields consistent improvements across a wide spectrum of IML models on multiple benchmarks. These ubiquitous gains underscore the efficacy of our approach in artifacts extraction and enhancing cross-domain generalization, independent of the specific network structure.

In summary, our contributions are threefold:


\begin{itemize}
    \item \textbf{Explicit Artifacts Modeling Paradigm}: We identify the latent nature of artifacts and reformulate IML as a latent variable learning problem, thus highlighting the necessity of explicitly modeling artifacts.

    \item \textbf{Two-stage Disentangled Learning Framework \& EditGroup-45K Dataset}: 
    We propose a PAL, which leverages the edit relation provided by our collected EditGroup-45K dataset, to disentangle the explicit model artifacts through two training stages.

    \item \textbf{Performance Boost and Validation}: 
    We demonstrate that our PAL yields consistent improvements across diverse IML architectures, and further conduct empirical analysis to validate that PAL does capture artifacts explicitly through feature disentanglement.

\end{itemize}

  \section{Related Works}

\subsection{Forgery Methods}

Deep forensic methods have progressively improved representation learning. MVSS-Net~\cite{mvsspp_2022} exploits boundary and noise cues, while PSCC-Net~\cite{liu2022pscc} models spatial-channel correlations. TruFor~\cite{trufor2023} integrates forensic and semantic features, CFM~\cite{luo2023beyond} mines critical forgery clues, IML-ViT~\cite{ma2023iml} adopts Transformer modeling, and NCL~\cite{NCL_IML_2023} handles noisy supervision.

Recent methods further diversify forensic modeling. APSC-Net~\cite{qu2024towards} improves manipulation localization, while FA-ViT~\cite{luo2024forgery} targets generalized face forgery detection. SparseViT~\cite{su2025spase} introduces sparse modeling, MoE-FFD~\cite{kong2025moe} employs mixture-of-experts, and RITA~\cite{zhu2026revisiting} reformulates localization as autoregressive sequence prediction. Mesorch~\cite{zhu2025mesorch} fuses semantic and noise cues, while DAF~\cite{qu2025revisiting} targets open-set tampered scene text detection.

More recently, Fake-HR1~\cite{fakehr1} explores visual reasoning, Omni-IML~\cite{qu2026omni} unifies interpretable localization, and Ivy-Fake~\cite{jiang2025ivy} enables explainable AIGC detection. TextShield-R1~\cite{qu2026textshield} employs reinforced reasoning, TADiff~\cite{bao2026activityforensics} targets activity-level video forgery localization, and OSDFD~\cite{kong2026open} addresses open-set deepfake detection. DS-Net~\cite{qu2026detect} detects AI-counterfeited text images, while Venus-DeFakerOne~\cite{DeFakerOne} unifies detection and localization.

Despite these advances, direct $P(y|x)$ learning can entangle artifacts with semantics, leading to \textit{Semantic Shortcuts}~\cite{yan2024effort,zhou2025beyond} and \textit{Label Noise}~\cite{NCL_IML_2023}.

To address this limitation, we introduce the PAL framework within a two-stage strategy. Instead of conventional end-to-end learning, we first explicitly approximate the latent artifacts $z$ to decouple them from semantic interference. Crucially, this learned latent space then serves as a superior initialization for downstream detectors, enabling standard architectures to break free from semantic shortcuts and achieve significant gains in generalization.

\subsection{Forgery Datasets}

Early benchmarks such as NIST16~\cite{NIST16_2019} and IMD20~\cite{IMD20_2020} provide real-world manipulations, while DEFACTO~\cite{defacto_2019} introduces large-scale synthetic forgeries. tampCOCO~\cite{CAT-Net2022} expands synthetic training data, while CIMD~\cite{zhang2024new} focuses on challenging manipulation cases. More recent datasets further improve scale and diversity, including MIML~\cite{qu2024towards}, OpenSDI~\cite{wang2025opensdi}, and OSTF~\cite{qu2025revisiting}. However, they lack explicit original-to-edit group structures required for pairwise artifact learning.

To address these limitations, we construct the \textbf{EditGroup-45K} dataset. It not only offers a large-scale volume of approximately 285K images but, more importantly, provides an explicit structure of 45K Edit Groups. This design enables us to leverage large-scale paired supervision signals for \textit{pre-finetuning}, facilitating the learning of robust and transferable artifacts features.
  \section{The EditGroup-45K Dataset}
\label{sec:dataset}

Existing IML benchmarks mainly collect manipulated images without explicit links to their authentic sources. As shown in Table~\ref{tab:dataset_comparison}, large-scale datasets often lack authentic anchors, while anchored datasets are usually small. To address this gap, we introduce \textbf{EditGroup-45K}, a large-scale dataset organized into source-anchored manipulation groups. This design supports provenance tracking at scale and provides the basis for relational artifacts modeling.

\subsection{Construction and Organization}
\label{subsec:construction}

\paragraph{Data Source and Generation Pipeline.}
The authentic anchor images in EditGroup-45K are curated from M$^3$~\cite{yang2025m}, considering image resolution and semantic diversity.
We follow the basic pre-processing and post-processing protocol of M$^3$ for quality and format consistency, while further constructing source-anchored edit groups with explicit edit relations.
Based on this protocol, we generate manipulated variants from two main sources:

\begin{itemize}
    \item \textit{Manual Manipulation:} We manually edit images using diverse editing tools, such as Adobe Photoshop, to simulate traditional tampering workflows. These manipulations cover common forensic categories, including copy-move, splicing, and removal.

    \item \textit{AIGC Synthesis:} We use advanced vision-language generation models, including Doubao~\cite{seedream2025seedream}, Qwen-Edit~\cite{wu2025qwenimagetechnicalreport}, Banana~\cite{team2023gemini}, and GPT-Image-1~\cite{hurst2024gpt}, to perform context-aware inpainting and content alteration.
\end{itemize}

\paragraph{Group Structure Definition.}
Based on these diverse generation methods, we organize the dataset into \textit{Edit Groups}. Unlike prior works that treat samples as independent instances, we formalize the dataset structure by explicitly anchoring each group to its authentic source(s):
\begin{equation}
\begin{aligned}
\mathcal{G}_i 
&= \Big\{ x_{anc}^{i,k} \Big\}_{k=1}^{M_i} 
   \cup \Big\{ x_{man}^{i,j} \Big\}_{j=1}^{N_i}, \\
\text{where} \quad 
&M_i \ge 1,\; N_i \ge 1.
\end{aligned}
\end{equation}
This organization enforces strict genealogical dependency: every manipulated image is physically linked to its specific source anchor(s). Consequently, while all samples within a group share the same underlying semantic content, they differ precisely in the applied edits. In total, EditGroup-45K consists of 45,546 aligned edit groups, comprising 52,074 authentic images and 232,520 manipulated variants. Detailed statistical distributions are provided in the Appendix~\ref{app:stats_details}.


\begin{wraptable}{L}{0.50\textwidth}
\vspace{-4mm}
\centering
\caption{Comparison with representative IML datasets.
`Anchor' denotes source availability.}
\label{tab:dataset_comparison}
\vspace{-2mm}

\scriptsize
\setlength{\tabcolsep}{2.6pt}
\renewcommand{\arraystretch}{0.90}

\resizebox{\linewidth}{!}{
\begin{tabular}{l|c|c|c|c}
\toprule
\textbf{Dataset} & \textbf{Auth.} & \textbf{Manip.} & \textbf{Anchor} & \textbf{Tech.} \\
\midrule
CASIA v2~\cite{CASIA_2013}              & 7,491  & 5,123   & \ding{55} & Manual \\
NIST'16~\cite{NIST16_2019}              & 0      & 564     & \ding{55} & Manual \\
IMD2020~\cite{IMD20_2020}               & 414    & 2,010   & \ding{55} & Manual \\
COVERAGE~\cite{Coverage_2016}            & 100    & 100     & \ding{51} & Manual \\
Columbia~\cite{Columbia_2006}            & 183    & 180     & \ding{51} & Manual \\
Fantastic Reality~\cite{Fantastic_Reality} & 16,592 & 19,423 & \ding{51} & Manual \\
DEFACTO~\cite{defacto_2019}       & 0      & 229,000 & \ding{55} & Manual \\
tampCOCO~\cite{CAT-Net2022}              & 0      & 800,000 & \ding{55} & Manual \\
MIML~\cite{qu2024towards}                & 0      & 123,150 & \ding{55} & Manual \\
CIMD~\cite{zhang2024new}                 & 400    & 400     & \ding{51} & Manual \\
\midrule
CocoGlide~\cite{trufor2023} & 512 & 512 & \ding{51} & AIGC \\
AutoSplice~\cite{jia2023autosplice} & 2,273 & 3,621 & \ding{51} & AIGC \\
\midrule
\rowcolor{gray!15}
\textbf{EditGroup-45K} & 52,074 & 232,520 & \ding{51} & \textbf{Hybrid} \\
\bottomrule
\end{tabular}
}

\vspace{-4mm}
\end{wraptable}

\subsection{Comparison with Existing Benchmarks}
\label{subsec:comparison}

Table~\ref{tab:dataset_comparison} compares EditGroup-45K with representative IML benchmarks. Existing datasets generally face two main limitations. First, regarding \textit{anchor availability}, current large-scale datasets (e.g., DEFACTO, tampCOCO, MIML) typically lack authentic anchors, making it impossible to model the relationship between original and edited images. While some datasets like Fantastic Reality and CIMD do provide anchors, they are often too small to support effective training. Second, regarding \textit{diversity}, prior works usually focus on only one type of manipulation: either traditional manual editing (e.g., Photoshop splicing in IMD2020 and CIMD) or, more recently, AI-generated synthesis (e.g., CocoGlide, AutoSplice). This separation fails to reflect the complexity of image manipulation, where both traditional and AI-based attacks coexist.

\textbf{EditGroup-45K} bridges these gaps. It is the first large-scale benchmark to offer simultaneously:
(1) explicit source-to-manipulation correspondence for group-wise modeling; and
(2) hybrid manipulation techniques, encompassing both traditional hand-crafted edits and emerging AIGC-based attacks.
Specifically, EditGroup-45K contains 140,137 manual edits (60.27\%) and 92,383 AIGC edits (39.73\%), covering copy-move (70,958), splicing (60,603), inpainting (50,911), and full-image generation (50,048).
To avoid shortcuts from non-manipulation-related low-level cues, both authentic anchors and manipulated images are processed with an identical post-processing pipeline.
This comprehensive coverage enables models trained on EditGroup-45K to learn generalizable artifact patterns across diverse manipulation sources.
  \begin{figure*}[t]
    \centering
    \includegraphics[width=0.975\textwidth]{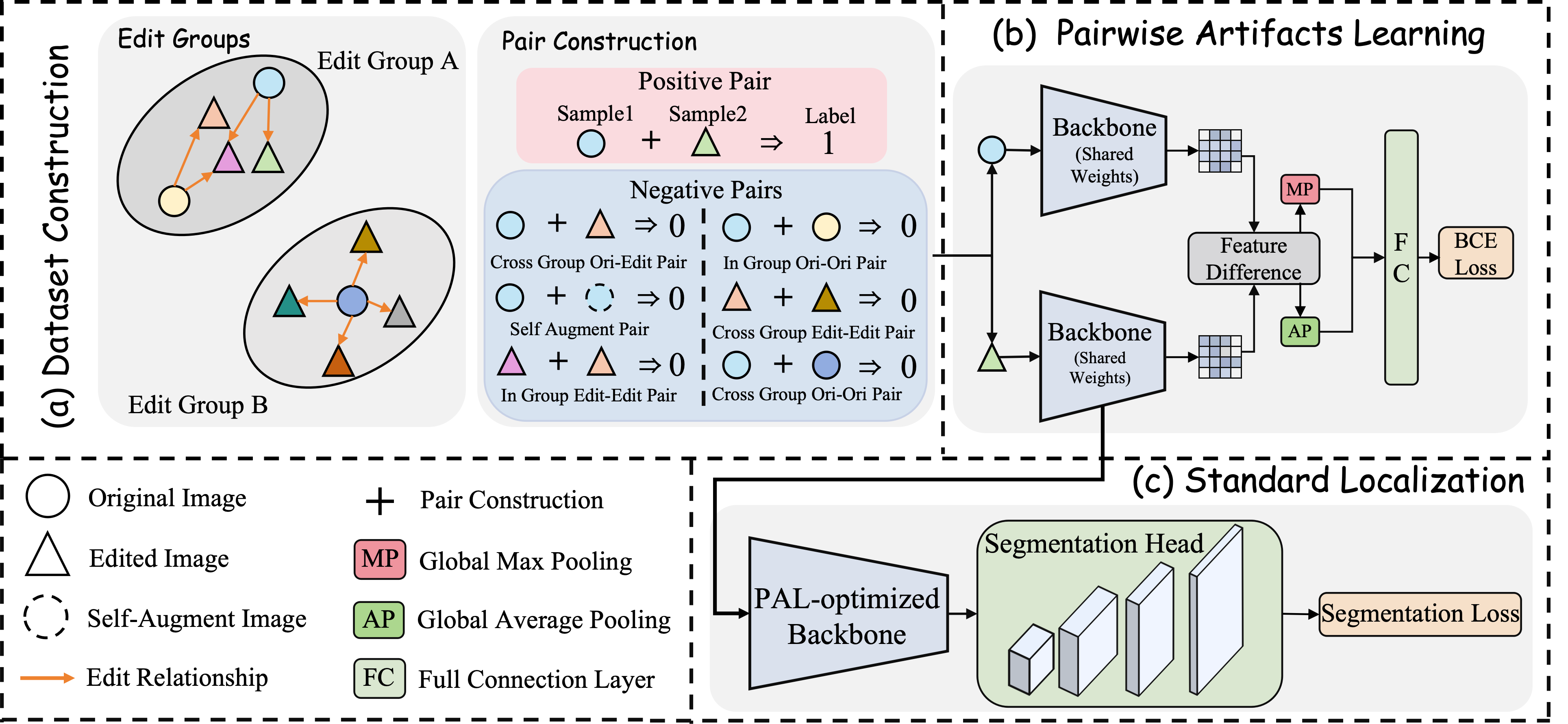}
    \caption{\textbf{Overview of the proposed two-stage paradigm.}
    (a) EditGroup-based training pair construction.
    (b) Pairwise Artifacts Learning (PAL) with a shared backbone and pairwise BCE supervision.
    (c) Standard Localization with the PAL-optimized backbone.}
    \label{fig:overview}
\end{figure*}

\section{Method}
\label{sec:method}

We propose a two-stage paradigm comprising \textit{Pairwise Artifacts Learning (PAL)} and \textit{Standard Localization (SL)} (Fig.~\ref{fig:overview}). 
In the first stage, we explicitly approximate the artifacts posterior $P(z|x)$. 
Subsequently, the PAL-optimized encoder serves as a robust initialization for the second stage, simplifying the optimization of standard IML models.

Our theoretical analysis shows that minimizing the pairwise BCE objective leads the PAL discriminator to estimate a conditional likelihood ratio between edit-related and non-edit-related pairs. 
Under the marginal matching condition, marginal semantic priors are suppressed, forcing the discriminator to rely on edit-induced artifacts. 
The full derivation and interpretation are provided in Appendix~\ref{app:theory}, with the optimal score and marginal matching condition given in Eq.~\eqref{eq:app_final_score} and Eq.~\eqref{eq:app_marginal_matching}, respectively.

Crucially, the success of the first stage hinges on the quality of relational supervision. 
To this end, we leverage the \textbf{EditGroup-45K} dataset to construct specific training pairs that enforce the theoretical \textbf{marginal matching constraint} in Eq.~\eqref{eq:app_marginal_matching}.

\subsection{Training Pairs Construction}
\label{subsec:pair_construction}

To explicitly guide artifact-oriented learning, the training pairs must align with the marginal matching condition in Eq.~\eqref{eq:app_marginal_matching}.
Specifically, positive pairs instantiate the manipulation mechanism $q_{\mathrm{rel}}$, while negative pairs approximate the marginal distribution $q_{\mathrm{rel}}(x_b)$ without sharing a direct edit relation with $x_a$.
Based on the group-structured EditGroup-45K dataset, we implement this via the following sampling strategies.

\paragraph{Positive Pairs.}
Positive samples are strictly defined as edit-related pairs $(x_a, x_b) \in \mathcal{P}^{+}$, where $x_b$ is a manipulated version directly derived from $x_a$. These pairs instantiate the \textbf{verified manipulation relationship} $q_{\mathrm{rel}}(x_b|x_a)$ defined in Eq.~\eqref{eq:app_theory_joint}, ensuring the model learns the specific inconsistencies introduced by the editing operation.

\paragraph{Negative Pairs (Marginal Matching).}
To enforce the marginal matching constraint ($q_{\mathrm{neg}}(x_b) \approx q_{\mathrm{rel}}(x_b)$), we construct negative pairs where $x_b$ exhibits the statistical characteristics of manipulated or diverse images but lacks a direct edit relationship with $x_a$. We design six distinct negative types to prevent the model from overfitting to semantic content:

\begin{itemize}
    \item \textbf{Cross-group Original-Edited}: An authentic image paired with a manipulated image from another group, discouraging shortcuts based on generic ``fake'' statistics.
    \item \textbf{Self-Augment}: An authentic image paired with its augmented view. This serves as a robust baseline, distinguishing forensic artifacts from benign content-preserving operations.
    \item \textbf{In-group Edited-Edited}: Two different manipulated images derived from the same source. This decouples the ``manipulated status" from the specific edit artifacts.
    \item \textbf{In-group Original-Original}: Two different authentic images from the same group. This prevents the model from using semantic similarity as a shortcut.
    \item \textbf{Global Edited-Edited}: Two manipulated images from different groups. This introduces global diversity, preventing overfitting to specific manipulation patterns.
    \item \textbf{Global Original-Original}: Two authentic images from different groups. This contrasts with purely semantic variations.
\end{itemize}

During training, for every positive pair, we sample negative pairs from this mixture. This sampling strategy follows the theoretical requirement discussed in Appendix~\eqref{app:marginal_matching}, forcing the discriminator to rely solely on the existence of a specific pairwise edit relationship.

\subsection{Stage I: Pairwise Artifacts Learning}
\label{subsec:stage1}
With the constructed training pairs, we employ the PAL (Fig.~\ref{fig:overview}b) to execute the disentanglement.

\paragraph{Shared Encoder \& Feature Difference.}
Let $f_{\theta}$ be a shared encoder. We compute the absolute difference $\mathbf{F}_{\Delta} = |f_{\theta}(x_a) - f_{\theta}(x_b)|$ and aggregate it via Global Average and Max Pooling to obtain the relation vector $\mathbf{z}$.

\paragraph{Optimization.}
We optimize the pairwise Binary Cross-Entropy (BCE) loss:
\begin{equation}
\label{eq:bce}
\mathcal{L}_{\text{PAL}} = \mathbb{E}_{((x_a,x_b),y)\sim\mathcal{D}} \left[ \mathrm{BCE}(g_{\phi}(\mathbf{z}), y) \right].
\end{equation}
Here, $g_{\phi}(\cdot)$ denotes a lightweight decoder that maps the latent representation $\mathbf{z}$ to a scalar relation score.
As demonstrated in Eq.~\eqref{eq:app_final_score}, minimizing this objective drives the scoring function to estimate the conditional likelihood ratio $s^* = \log(q_{\mathrm{rel}}/q_{\mathrm{neg}})$. Given that our data construction satisfies the marginal matching constraint, this optimization effectively compels the encoder to emphasize artifact-related signals while suppressing semantic content.

\subsection{Stage II: Standard Localization}
\label{subsec:stage2}

In this stage (Fig.~\ref{fig:overview}c), we initialize the backbone of target IML models with the PAL-optimized encoder learned in Stage~I.
The subsequent training strictly follows standard pixel-level localization protocols~\cite{ma2025imdl}.
This design effectively reduces the IML objective from the entangled conditional distribution $P(y|x)$ to the cleaner conditional distribution $P(y|z)$,
thereby validating our core theoretical insight that manipulation artifacts can be explicitly modeled and transferred.

  \section{Experiments}
This section provides a comprehensive evaluation of the proposed Pairwise Artifacts Learning (PAL) stage.
We first assess the overall effectiveness of PAL under the standard localization setting (Sec.~\ref{subsec:ablation}).
We then conduct detailed ablation studies to analyze the contribution of different supervision forms, pair representations, and negative pair compositions (Sec.~\ref{subsec:ablation_supervision}–\ref{subsec:ablation_negative}).
Furthermore, we examine how PAL approximates manipulation artifacts through quantitative boundary analysis and qualitative visualization (Sec.~\ref{subsec:qualitative_artifact}).
Finally, we evaluate the transferability of PAL across diverse backbone architectures and existing IML baselines to verify its general applicability (Sec.~\ref{subsec:transferability}).


\begin{table}[t]
\centering

\begin{minipage}[t]{0.49\textwidth}
\centering
\captionof{table}{Ablation study summary. \textbf{In-Avg} denotes in-domain performance on CASIA v1. \textbf{Cross-Avg} measures cross-domain generalization. \textbf{All-Avg} reports the mean over all six datasets. Detailed results are in Appendix~\ref{app:ablation}.}
\label{tab:PAL_ablation_summary}
\vspace{1mm}
\resizebox{\linewidth}{!}{
\begin{tabular}{lccc}
\toprule
\textbf{Model} & \textbf{In-Avg} & \textbf{Cross-Avg} & \textbf{All-Avg} \\
\midrule
Baseline & 0.7369 & 0.5320 & 0.5661 \\
Baseline-ABS & 0.8457 & 0.5292 & 0.5820 \\
\midrule
PAL (Ours) & 0.8256 & 0.5868 & 0.6266 \\
\midrule
w/o Diff & 0.8073 & 0.5333 & 0.5790 \\
w/o AvgPool & 0.7814 & 0.5438 & 0.5834 \\
w/o MaxPool & 0.8368 & 0.5676 & 0.6125 \\
\midrule
w/o CG-OE & 0.8232 & 0.5523 & 0.5975 \\
w/o SA & 0.8521 & 0.5492 & 0.5997 \\
w/o IG-EE & 0.7893 & 0.5386 & 0.5804 \\
w/o IG-OO & 0.8239 & 0.5539 & 0.5989 \\
w/o G-EE & 0.7790 & 0.5511 & 0.5891 \\
w/o G-OO & 0.8183 & 0.5700 & 0.6114 \\
\bottomrule
\end{tabular}}
\end{minipage}
\hfill
\begin{minipage}[t]{0.48\textwidth}
\centering

\captionof{table}{Boundary activation strength before and after training across datasets.}
\label{tab:boundary_activation}
\vspace{1mm}
\resizebox{0.95\linewidth}{!}{
\begin{tabular}{lcccc}
\toprule
Dataset & ImageNet & PAL & Increase & Ratio \\
\midrule
CASIAv1    & 0.48 & 3.03 & 2.55 & 6.32 \\
Columbia  & 0.47 & 2.36 & 1.89 & 5.00 \\
Coverage  & 0.46 & 3.38 & 2.92 & 7.40 \\
NIST16    & 0.58 & 3.24 & 2.66 & 5.62 \\
AutoSplice& 0.41 & 2.74 & 2.33 & 6.64 \\
CocoGlide & 0.46 & 2.83 & 2.38 & 6.22 \\
\bottomrule
\end{tabular}}

\vspace{4mm}

\captionof{table}{Transferability of PAL initialization. 
}
\label{tab:backbone_transfer}
\vspace{-2mm}
\resizebox{\linewidth}{!}{
\begin{tabular}{lccc}
\toprule
\textbf{Model} 
& \textbf{Standard} 
& \textbf{PAL} 
& $\boldsymbol{\Delta_{rel}}$ (\%) \\
\midrule
ResNet-101        & 0.4878 & 0.5226 & +7.12 \\
SegFormer-B3      & 0.5496 & 0.5949 & +8.24 \\
Swin-Small        & 0.5440 & 0.5781 & +6.27 \\ 
ConvNeXt-Small    & 0.5661 & 0.6266 & +10.68 \\
DINOv3-Base & 0.6114 & 0.6477 & +5.9 \\
SAM-Base    & 0.5206 & 0.5937 & +14.0 \\
\midrule
CAT-Net           & 0.5452 & 0.5812 & +6.60 \\
PSCC              & 0.5281 & 0.5583 & +5.72 \\
MVSS              & 0.4946 & 0.5437 & +9.93 \\
Mesorch           & 0.5509 & 0.5853 & +6.24 \\
TruFor            & 0.5623 & 0.6095 & +8.39 \\
\bottomrule
\end{tabular}}

\end{minipage}
\vspace{-6mm}
\end{table}
\subsection{Experimental Setup}

\paragraph{Stage I: PAL.}
We employ the PAL stage as the foundation before SL, training a shared encoder with the binary objective in Eq.~\eqref{eq:bce} on edit-related and non-edit-related pairs constructed as in Sec.~\ref{subsec:pair_construction}.

\paragraph{Stage II: SL.}
The subsequent localization training (referred to as \textbf{Stage II}) follows the Protocol-CAT setting~\cite{ma2025imdl}.
The primary training datasets include CASIA v2~\cite{CASIA_2013}, IMD2020~\cite{IMD20_2020}, FantasticReality~\cite{Fantastic_Reality}, and Tampered COCO~\cite{CAT-Net2022}.
Evaluation is conducted on CASIA v1~\cite{CASIA_2013}, Coverage~\cite{Coverage_2016}, NIST16~\cite{NIST16_2019}, Columbia~\cite{Columbia_2006}, AutoSplice~\cite{jia2023autosplice}, and COCO-Glide~\cite{trufor2023}.

\paragraph{Metrics.}
Unless otherwise specified, all results report the binary pixel-level F1 score~\cite{ma2025imdl} of the Stage II model, with masks binarized at a threshold of 0.5.

\subsection{Ablation Study}
\label{subsec:ablation}

We conduct an ablation study to evaluate the effectiveness of the PAL stage and its key design components, with all metrics reported on the final SL task under the Protocol-CAT setting.
To ensure fair comparison, all variants use the same ConvNeXt-Small backbone (pretrained on ImageNet-1K) and undergo identical SL training, differing only in the supervision strategies or design choices within the PAL stage.
All results are summarized in Table~\ref{tab:PAL_ablation_summary}.

\subsubsection{Effect of Pairwise Artifacts Learning}
\label{subsec:ablation_supervision}

We compare different supervision forms: \textbf{Baseline} is initialized with ImageNet weights only. 
\textbf{Baseline-ABS} applies absolute binary supervision on EditGroup-45K, where authentic images are labeled as 0 and edited images as 1, without exploiting pairwise relations. 
\textbf{PAL} employs our proposed pairwise learning strategy to achieve artifact disentanglement.

As shown in Table~\ref{tab:PAL_ablation_summary}, PAL achieves the highest All-Avg of 0.6266.
While Baseline-ABS excels in In-Avg, its Cross-Avg (0.5292) falls slightly below the Baseline (0.5320), suggesting overfitting to the source domain.
In contrast, PAL significantly improves Cross-Avg to 0.5868.
This demonstrates that while absolute supervision is prone to overfitting, our pairwise modeling successfully learns robust, transferable representations for unseen manipulation patterns.

\subsubsection{Effect of Pair Representation Design}
\label{subsec:ablation_representation}

We evaluate the design of pair representation within the PAL: \textbf{w/o Diff} removes explicit feature difference modeling; \textbf{w/o AvgPool} and \textbf{w/o MaxPool} exclude global average and max pooling, respectively.

As shown in Table~\ref{tab:PAL_ablation_summary}, removing explicit difference modeling (\textbf{w/o Diff}) results in the most significant drop in \textbf{Cross-Avg} (from 0.5868 to 0.5333). This confirms that the explicit difference map is the cornerstone for extracting generalized manipulation traces.
Regarding pooling strategies, \textbf{w/o AvgPool} suffers the lowest \textbf{In-Avg} (0.7814), indicating that global context aggregation is essential for stability in the source domain.
Conversely, while \textbf{w/o MaxPool} maintains high in-domain performance, its \textbf{Cross-Avg} decreases to 0.5676, suggesting that capturing peak local inconsistencies is vital for cross-domain robustness.
These results demonstrate that difference modeling, combined with complementary pooling strategies, is crucial for balancing source reliability and generalization.

\subsubsection{Effect of Negative Pair Composition}
\label{subsec:ablation_negative}

We analyze the impact of specific negative pair types, denoting Original as `O' and Edited as `E'.
As shown in Table~\ref{tab:PAL_ablation_summary}, removing most negative components leads to a consistent degradation in both In-Domain and Cross-Domain performance.
However, distinct behaviors are observed for Self-Augmented negatives and In-Group pairs.

Most notably, removing Self-Augmented negatives (\textbf{w/o SA}) results in a divergent trend: In-Avg spikes to the highest value of 0.8521 (surpassing PAL), while Cross-Avg significantly drops to 0.5492.
This sharp discrepancy indicates that without the constraint of self-augmentation, the model learns trivial shortcuts specific to the source domain (overfitting), thereby achieving high training scores but sacrificing robustness to unseen artifacts.

In contrast, removing In-group Edited--Edited pairs (\textbf{w/o IG-EE}) causes the most severe decline in generalization, yielding the lowest Cross-Avg of 0.5386.
This confirms that distinguishing between different manipulation types within the same semantic context provides the tightest constraint.
Essentially, IG-EE defines the most critical marginal matching, forcing the model to decouple artifacts from semantics, which is the key driver for the stage's cross-domain success.

\subsubsection{BCE vs. Contrastive Objective}
We further compare BCE with a contrastive variant to justify the loss design.
BCE is better aligned with EditGroup-45K, where manipulation artifacts are heterogeneous and do not naturally satisfy the class-level compactness assumption required by contrastive pulling.
Empirically, BCE leads to faster convergence and a higher performance ceiling than the contrastive objective.
Detailed analysis is provided in Appendix~\ref{app:bce_contrastive}.


\subsection{Analysis of Artifacts Approximation}
\label{subsec:qualitative_artifact}

\paragraph{Quantitative Analysis.}
Since manipulation artifacts often concentrate near boundaries~\cite{mvsspp_2022}, boundary responses serve as a natural probe for artifact sensitivity.
We therefore compare an ImageNet-pretrained ConvNeXt-Small backbone with its PAL-optimized counterpart by measuring feature activations along boundaries extracted via $7\times7$ morphological erosion~\cite{ma2023iml}.

As shown in Table~\ref{tab:boundary_activation}, PAL consistently amplifies boundary activations by approximately $5\times$–$7\times$ across all datasets, indicating enhanced sensitivity to manipulation artifacts.

\begin{wrapfigure}[17]{r}{0.495\textwidth}
\vspace{-4mm}
\centering
\includegraphics[width=\linewidth]{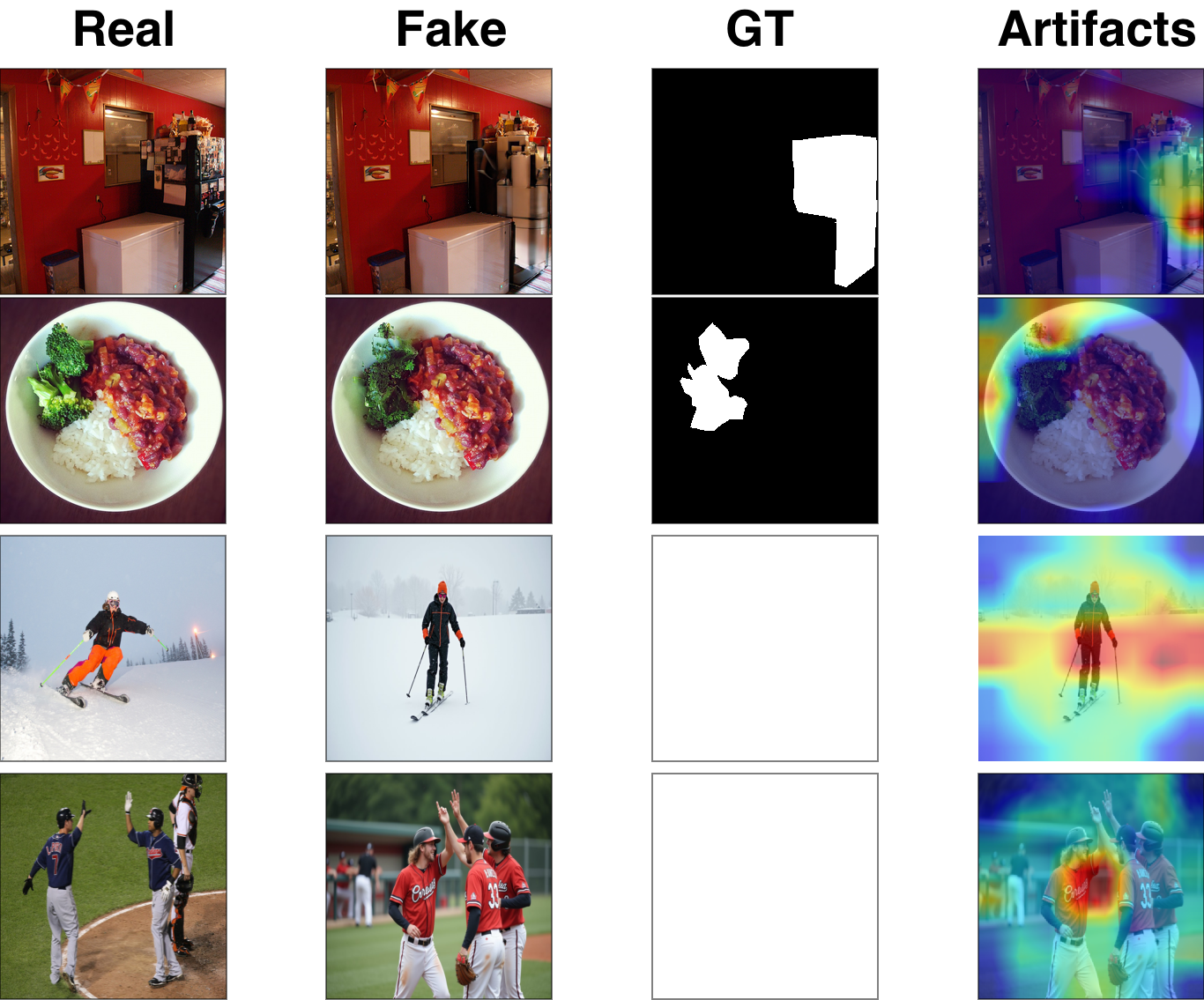}
\vspace{-1mm}
\caption{Qualitative visualization of PAL-derived Artifacts.}
\label{fig:artifact_vis}
\end{wrapfigure}

\paragraph{Qualitative Analysis.}
We visualize the \textbf{Artifacts} responses derived from PAL to examine what information is captured beyond semantic changes.
Following the derivation in Appendix~\ref{app:artifact_derivation}, PAL first computes pairwise feature differences between real and manipulated images; the semantic component is then suppressed to obtain residual artifact responses.
As shown in Fig.~\ref{fig:artifact_vis}, the first two rows correspond to manual local manipulations, which mainly activate manipulated boundaries, while the last two rows correspond to AIGC-based full-image edits, which produce broader responses.
This suggests that PAL captures manipulation-dependent artifacts rather than a uniform boundary cue.

\subsection{Transferability Analysis}
\label{subsec:transferability}

We evaluate the transferability of the PAL stage across diverse backbone architectures.
All experiments follow the SL setting under the Protocol-CAT benchmark.
The goal is to verify that the effectiveness of PAL is architecture-agnostic and not limited to a specific backbone design.

\subsubsection{Transferability Across Backbones}
\label{subsec:transfer_backbone}

We evaluate the transferability of PAL across diverse backbone families, including CNNs, Transformers, and recent foundation-style vision backbones. 
Specifically, we consider ResNet-101~\cite{Resnet_2016}, ConvNeXt-Small~\cite{Convnet_2022}, Swin-Small~\cite{Swin_2021}, SegFormer-B3~\cite{SegFormer_2021}, DINOv3-Base~\cite{simeoni2025dinov3}, and SAM-Base~\cite{kirillov2023sam}. 
These models cover standard convolutional architectures, modern ConvNet designs, hierarchical vision Transformers, semantic segmentation backbones, self-supervised visual representations, and promptable segmentation backbones.

As shown in the top section of Table~\ref{tab:backbone_transfer}, PAL initialization consistently improves over standard pretraining across all tested backbones. 
For conventional backbones, PAL brings relative gains from 6.27\% to 10.68\%, with ConvNeXt-Small achieving the largest improvement among them (+10.68\%). 
Transformer-based models also benefit, with SegFormer-B3 and Swin-Small improving by 8.24\% and 6.27\%, respectively. 
Moreover, PAL remains effective when combined with stronger modern backbones: DINOv3-Base improves from 0.6114 to 0.6477 (+5.94\%), and SAM-Base improves from 0.5206 to 0.5937 (+14.04\%). 
These results demonstrate that PAL provides an architecture-agnostic artifact-aware initialization, rather than relying on convolution-specific inductive biases.

\subsubsection{Transferability to Baseline Methods}
\label{subsec:transfer_baseline}

We further evaluate the transferability of PAL by replacing the backbone initializations within existing baseline methods, including CAT-Net~\cite{CAT-Net2022}, PSCC~\cite{liu2022pscc}, MVSS~\cite{mvsspp_2022}, Mesorch~\cite{zhu2025mesorch}, and TruFor~\cite{trufor2023}.
In this setting, we maintain the original model architectures and training protocols, substituting only their ImageNet-pretrained backbones with PAL-optimized weights.

As summarized in Table~\ref{tab:backbone_transfer} bottom, PAL stage consistently boosts performance across all evaluated baselines.
Notably, TruFor achieves a substantial relative gain of 8.39\% (0.5623 $\to$ 0.6095), while Mesorch improves by 6.24\%.
Furthermore, earlier methods like MVSS observe the largest improvement of 9.93\%.
These results indicate that PAL learns generic, transferable representations that can be seamlessly injected into diverse image manipulation localization frameworks to significantly enhance their performance.

  \section{Conclusion}
In this work, we validate the hypothesis that IML is most effectively formulated as a latent-variable learning problem, governed by the probabilistic decomposition: $P(y|x) = \int P(y|z)P(z|x)dz$. By shifting away from the prevailing end-to-end implicit learning paradigm, we demonstrate that the ``invisible" artifacts can be explicitly disentangled from semantic content through relational contrast, without pixel-level artifact annotation. The deployment of the Pairwise Artifacts Learning (PAL) stage, supported by the EditGroup-45K dataset, confirms that modeling the posterior $P(z|x)$ is not only theoretically feasible but also practically superior to the implicit paradigm.

Furthermore, the universal performance gains observed across diverse models indicate that our two-stage paradigm offers a robust solution to the generalization bottlenecks caused by semantic shortcuts. By decoupling artifact extraction from the final localization task, we provide a generic and effective initialization strategy that transcends specific network architectures. Ultimately, this research establishes a new direction for image forensics: moving beyond reliance on intrinsic network capacity toward data-driven, explicit artifacts representation.

  \bibliography{bibliography}
  \bibliographystyle{plain}

  \newpage
  \appendix

\section{Theoretical Analysis}
\label{app:theory}

This appendix provides the theoretical justification for PAL. We first derive the Bayes-optimal solution of the pairwise BCE objective, and then interpret why marginally matched negative sampling encourages artifact-focused discrimination.

\subsection{Bayes-Optimal Solution of the BCE Objective}
\label{app:bayes_opt}

\paragraph{Pairwise Discrimination Setup.}
Let $(x_a, x_b)$ be an image pair and $y\in\{0,1\}$ indicate whether $x_b$ is directly edited from $x_a$.
We define the positive and negative pair distributions as
\begin{equation}
\begin{aligned}
    p_{+}(x_a,x_b)=p(x_a)\,q_{\mathrm{rel}}(x_b\mid x_a), \\
    p_{-}(x_a,x_b)=p(x_a)\,q_{\mathrm{neg}}(x_b\mid x_a),
\end{aligned}
\label{eq:app_theory_joint}
\end{equation}
where $q_{\mathrm{rel}}$ denotes the conditional manipulation mechanism that introduces edit-induced artifacts, and $q_{\mathrm{neg}}$ denotes the non-related sampling strategy.

\paragraph{Population BCE Risk.}
Let $s(x_a,x_b)\in\mathbb{R}$ be the scoring function and $p=\sigma(s)\in(0,1)$ be the predicted probability of $y=1$.
The population BCE risk used in PAL is
\begin{equation}
\begin{aligned}
\mathcal{L}[s]
=
&-\mathbb{E}_{(x_a,x_b)\sim p_{+}}
  \big[\log \sigma(s)\big] \\
&-\mathbb{E}_{(x_a,x_b)\sim p_{-}}
  \big[\log(1-\sigma(s))\big].
\end{aligned}
\label{eq:app_bce_risk}
\end{equation}

\paragraph{Pointwise Minimization and Optimal Posterior.}
For a fixed pair $(x_a,x_b)$, denote $u=p_{+}(x_a,x_b)$ and $v=p_{-}(x_a,x_b)$.
Optimizing $\mathcal{L}$ reduces to the pointwise minimization of
\begin{equation}
\ell(p) = -u\log p - v\log(1-p),
\label{eq:app_pointwise}
\end{equation}
with respect to $p\in(0,1)$.
Setting $\partial \ell/\partial p = 0$ gives
\begin{equation}
p^{*}
=
\frac{u}{u+v}
=
\frac{p_{+}(x_a,x_b)}{p_{+}(x_a,x_b)+p_{-}(x_a,x_b)}.
\label{eq:app_opt_posterior}
\end{equation}

\paragraph{Optimal Log-Odds Scoring Function.}
The Bayes-optimal score is the log-odds of $p^{*}$:
\begin{equation}
s^{*}(x_a,x_b)
=
\log\frac{p^{*}}{1-p^{*}}
=
\log\frac{p_{+}(x_a,x_b)}{p_{-}(x_a,x_b)}.
\label{eq:app_opt_logodds}
\end{equation}

\paragraph{Reduction to a Conditional Likelihood Ratio.}
Using the factorization in Eq.~\eqref{eq:app_theory_joint}, we have
\begin{equation}
\frac{p_{+}(x_a,x_b)}{p_{-}(x_a,x_b)}
=
\frac{p(x_a)\,q_{\mathrm{rel}}(x_b|x_a)}
     {p(x_a)\,q_{\mathrm{neg}}(x_b|x_a)}
=
\frac{q_{\mathrm{rel}}(x_b|x_a)}
     {q_{\mathrm{neg}}(x_b|x_a)}.
\label{eq:app_cond_ratio}
\end{equation}
Substituting Eq.~\eqref{eq:app_cond_ratio} into Eq.~\eqref{eq:app_opt_logodds} yields
\begin{equation}
s^{*}(x_a,x_b)
=
\log\frac{q_{\mathrm{rel}}(x_b|x_a)}
         {q_{\mathrm{neg}}(x_b|x_a)}.
\label{eq:app_final_score}
\end{equation}

\subsection{Interpretation of Marginal Matching and Artifact-Focused Discrimination}
\label{app:marginal_matching}

\paragraph{Marginal Matching Condition.}
To prevent semantic shortcuts, we construct negative samples to satisfy
\begin{equation}
\begin{aligned}
q_{\mathrm{neg}}(x_b)\approx q_{\mathrm{rel}}(x_b), \\
\text{s.t. } q_{\mathrm{neg}}(x_b\mid x_a)\neq q_{\mathrm{rel}}(x_b\mid x_a).
\end{aligned}
\label{eq:app_marginal_matching}
\end{equation}

\paragraph{Potential Semantic Shortcut without Matching.}
If $q_{\mathrm{neg}}(x_b)$ differs significantly from $q_{\mathrm{rel}}(x_b)$, then $x_b$ alone may carry label information, e.g., negative samples being systematically cleaner or from different content domains.
In this case, the discriminator can reduce BCE loss by exploiting marginal semantic priors without modeling the edit relation between $(x_a,x_b)$.

\paragraph{Effect of Marginal Matching.}
By enforcing $q_{\mathrm{neg}}(x_b) \approx q_{\mathrm{rel}}(x_b)$, label information obtainable from the marginal distribution of $x_b$ is largely removed.
Therefore, under the Bayes-optimal objective in Eq.~\eqref{eq:app_final_score}, the discriminator must rely on the conditional discrepancy between $q_{\mathrm{rel}}(x_b|x_a)$ and $q_{\mathrm{neg}}(x_b|x_a)$.

\paragraph{Connection to Manipulation Artifacts.}
Since $q_{\mathrm{rel}}$ differs from $q_{\mathrm{neg}}$ due to the causal editing process, the remaining discriminative cue corresponds to edit-induced traces, i.e., latent artifacts.
Thus, PAL is theoretically encouraged to learn artifact-relevant representations rather than global semantic content.

\section{BCE vs. Contrastive Objective}
\label{app:bce_contrastive}

We further analyze the choice of the pairwise objective in PAL by comparing the proposed BCE objective with a contrastive variant, denoted as PAL-Con.
For a fair comparison, PAL-Con uses the same pair construction, shared encoder, feature difference operation, and pooling design as PAL, but replaces the binary classifier with a supervised contrastive objective over pair-level relation embeddings.

\paragraph{Contrastive Variant.}
Given an image pair $(x_a,x_b)$, we first compute the pairwise feature difference
\begin{equation}
\mathbf{F}_{\Delta}=|f_{\theta}(x_a)-f_{\theta}(x_b)|.
\end{equation}
After global average/max pooling, the relation representation is projected and normalized as
\begin{equation}
\mathbf{h}_i=\mathrm{norm}\big(q_{\phi}(\mathbf{z}_i)\big),
\end{equation}
where $\mathbf{z}_i$ denotes the pooled relation representation of the $i$-th pair.
PAL-Con then applies a supervised contrastive loss:
\begin{equation}
\mathcal{L}_{\mathrm{con}}
=
\sum_{i\in\mathcal{B}}
-\frac{1}{|\mathcal{P}(i)|}
\sum_{p\in\mathcal{P}(i)}
\log
\frac{
\exp(\mathrm{sim}(\mathbf{h}_i,\mathbf{h}_p)/\tau)
}{
\sum_{a\in\mathcal{B},a\neq i}
\exp(\mathrm{sim}(\mathbf{h}_i,\mathbf{h}_a)/\tau)
},
\end{equation}
where $\mathcal{P}(i)=\{p\in\mathcal{B}:y_p=y_i,p\neq i\}$, $\mathrm{sim}(\cdot,\cdot)$ denotes cosine similarity, and $\tau$ is the temperature.

\paragraph{Analysis.}
Table~\ref{tab:app_bce_contrastive} compares BCE-based PAL with the contrastive variant PAL-Con.
PAL-Con requires more training epochs and obtains lower downstream performance, while BCE-based PAL converges faster and reaches a higher performance ceiling.

This phenomenon is consistent with the data structure of EditGroup-45K.
Contrastive learning assumes that samples with the same label should be pulled together in the embedding space, while samples with different labels should be pushed apart.
However, manipulation artifacts are highly heterogeneous, covering copy-move, splicing, inpainting, and full-image generation.
Thus, edit-related pairs do not naturally form a compact class in the relation space, and forcing them to be close may introduce gradient conflicts among different manipulation patterns.

In contrast, BCE only requires discriminating edit-related from non-edit-related pairs, without imposing class-level compactness.
Therefore, BCE better matches the relational supervision in EditGroup-45K and is more suitable for PAL.

\begin{table}[t]
\centering
\caption{Comparison between BCE and contrastive objectives.
EditGroup-45K provides reliable inter-class differences but does not assume intra-class compactness among heterogeneous manipulation artifacts.
PAL-Con converges more slowly and reaches a lower performance ceiling than BCE-based PAL.}
\label{tab:app_bce_contrastive}
\resizebox{\linewidth}{!}{
\begin{tabular}{lcccccc}
\toprule
\textbf{Setting} 
& \textbf{Intra-class Compactness} 
& \textbf{Inter-class Difference} 
& \textbf{Epochs} 
& \textbf{In-Avg} 
& \textbf{Cross-Avg} 
& \textbf{All-Avg} \\
\midrule
EditGroup-45K Property & No & Yes & -- & -- & -- & -- \\
PAL-Con (Contrastive)  & Required & Required & 30 & 0.7953 & 0.5548 & 0.5951 \\
PAL (BCE)              & Not Required & Required & 12 & 0.8256 & 0.5868 & 0.6266 \\
\bottomrule
\end{tabular}}
\end{table}

Overall, BCE is more suitable for PAL because it matches the heterogeneous nature of manipulation artifacts, converges faster, and achieves better downstream localization performance.

\section{Derivation and Interpretation of Artifacts Approximation}
\label{app:artifact_derivation}

In this appendix, we provide the theoretical grounding for the visualization in Sec.~\ref{subsec:qualitative_artifact}, linking the feature-level responses back to the Bayes-optimal analysis in Sec.~\ref{app:theory}. We clarify why a residual subtraction is necessary to reveal the learned artifacts despite the theoretical guarantee of artifact disentanglement.

\paragraph{Feature Decomposition.}
Consider a specific positive pair instance consisting of a real image $r$ (acting as anchor $x_a$) and its manipulated counterpart $f$ (acting as $x_b$). The manipulated image $f$ is generated via a conditional mechanism $q_{\mathrm{rel}}(f|r)$ that introduces both explicit semantic edits (denoted as $E$) and latent manipulation artifacts ($z$).
Viewing the deep encoder as an information container, the feature representation $\phi(f)$ can be physically decomposed as:
\begin{equation}
\phi(f) \approx \phi_{\text{sem}}(r) + \phi_{\text{sem}}(E) + \phi_{\text{art}}(z),
\end{equation}
where $\phi_{\text{sem}}(\cdot)$ represents the semantic content shared with standard vision tasks (inherited from ImageNet pretraining), and $\phi_{\text{art}}(z)$ represents the manipulation-specific footprint derived from $q_{\mathrm{rel}}$.

\paragraph{The Gap between Optimal Score and Physical Representation.}
In Sec.~\ref{app:theory}, we proved that under the \textbf{Marginal Matching constraint} (Eq.~\eqref{eq:app_marginal_matching}), the optimal scoring function $s^*(r, f)$ reduces to the conditional likelihood ratio.
This implies that the \emph{decision boundary} is optimized to be orthogonal to semantic variations, focusing solely on the discrepancy caused by artifacts $z$.
\textbf{However, the optimization of the scoring function $s$ does not equate to the erasure of semantic information from the feature encoder $\phi$.}
Two practical factors bridge this gap:
\begin{itemize}
    \item \textbf{Model Bias:} Since the backbone retains strong priors from ImageNet initialization, it acts as a semantic filter by default, preserving content information ($\phi_{\text{sem}}$) in the spatial feature maps.
    \item \textbf{Data Reality:} A manipulated image $f$ cannot be infinitely close to $r$ in the semantic space; effective manipulation inherently requires semantic changes~\cite{zhu2025mesorch} (e.g., splicing a new object) to deceive human viewers. Thus, the term $\phi_{\text{sem}}(E)$ is physically inevitable.
\end{itemize}
Consequently, the raw pairwise difference $\mathbf{D}(r,f) = \|\phi(r) - \phi(f)\|$ captures the total variation, mixing both the inevitable semantic edit $\phi_{\text{sem}}(E)$ and the target artifacts $\phi_{\text{art}}(z)$.

\paragraph{Isolating the Artifact Signal.}
To strictly validate that our framework has aligned with the theoretical goal of modeling $q_{\mathrm{rel}}(f|r)$, we employ a residual probing mechanism.
We use a fixed ImageNet-pretrained encoder to approximate the pure semantic variance $\mathbf{S}(f) \approx \|\phi_{\text{sem}}(E)\|$.
By subtracting this semantic baseline from our model's response, we isolate the artifact component:
\begin{equation}
\mathbf{A}_{\text{vis}} = \max\big(\mathbf{D}(r,f) - \mathbf{S}(f),\,0\big) \approx \phi_{\text{art}}(z).
\end{equation}
This operation empirically verifies our theoretical derivation: while the raw features are dominated by the semantic bias of the backbone architecture, the PAL objective successfully forces the encoder to learn a structured signal $\phi_{\text{art}}(z)$ that exists specifically to satisfy the conditional ratio requirement.

\section{Hyperparameter Sensitivity Analysis}
\label{sec:appendix_hyperparams}

In this section, we present a detailed ablation study concerning the hyperparameters used in the Pairwise Artifacts Learning (PAL) framework.

\paragraph{Experimental Setup.}
We implemented our framework using the ForensicHub~\cite{du2025forensichubunifiedbenchmark} training framework. The experiments were conducted on a server equipped with 8 NVIDIA A100 GPUs. All input images were resized to $256 \times 256$ pixels during both training and inference stages.  For the experiments, the EditGroup-45K dataset was partitioned into a training set (95\%) and a hold-out testing set (5\%), with the split performed at the group level to prevent data leakage between original sources. In the testing phase, both positive and negative pairs were pre-sampled and fixed to ensure a consistent evaluation benchmark across all model variations. We employ the Area Under the Receiver Operating Characteristic Curve (AUC) as the primary evaluation metric to assess the model's discriminative capability in capturing latent artifacts $z$.

Regarding the negative sampling strategy, unless otherwise specified, the negative pairs are sampled dynamically from the six pools defined in Sec.~\ref{app:ablation_negative} according to the following probability distribution:
\textbf{Self-Augmented (SA)}: 0.2,
\textbf{In-group Edited-Edited (IG-EE)}: 0.2,
\textbf{In-group Original-Original (IG-OO)}: 0.2,
\textbf{Global Edited-Edited (G-EE)}: 0.15,
\textbf{Global Original-Original (G-OO)}: 0.15,
and \textbf{Cross-group Original-Edited (CG-OE)}: 0.1.
This distribution ensures a diverse coverage of negative types during the optimization process.
The ablation follows a sequential optimization strategy: we first determine the optimal learning rate, followed by the training duration (epochs), and finally the positive-to-negative sampling ratio.

\subsection{Impact of Learning Rate}
We first analyze the impact of the learning rate (LR) on model convergence. We fixed the epoch count to 12 and the sampling ratio to $1:1$. The learning rate was swept across a range from $2 \times 10^{-4}$ to $1 \times 10^{-6}$. As shown in Table~\ref{tab:ablation_lr}, the model achieves optimal performance at an LR of $1 \times 10^{-4}$. Increasing the LR to $2 \times 10^{-4}$ degrades performance to $0.8873$, likely due to optimization instability. Conversely, reducing the LR to $1 \times 10^{-6}$ results in a suboptimal AUC of $0.8862$, indicating underfitting or slow convergence. Consequently, we fix the learning rate at $1 \times 10^{-4}$ for subsequent experiments.

\begin{table}[h]
    \centering
    \caption{\textbf{Effect of Learning Rate.} The model is trained for 12 epochs with a 1:1 sampling ratio. The best performance is observed at $1 \times 10^{-4}$.}
    \label{tab:ablation_lr}
    \begin{tabular}{lc}
        \toprule
        \textbf{Learning Rate} & \textbf{Test AUC} \\
        \midrule
        $2 \times 10^{-4}$ & 0.8873 \\
        $\mathbf{1 \times 10^{-4}}$ & $\mathbf{0.8926}$ \\
        $2 \times 10^{-5}$ & 0.8897 \\
        $1 \times 10^{-6}$ & 0.8862 \\
        \bottomrule
    \end{tabular}
\end{table}

\subsection{Impact of Training Duration}
We further investigate the optimal training duration. We define one epoch as a complete traversal of all available positive pairs in the training set (100\% sampling), with negative pairs sampled dynamically. Table~\ref{tab:ablation_epoch} illustrates the AUC variations across different epochs. The model performance peaks at epoch 12 with an AUC of $0.8926$. At 10 epochs, the model has not fully converged ($0.8747$), while extending training to 15 epochs leads to a slight performance drop ($0.8920$), suggesting the onset of overfitting to the training pairs. Therefore, we select 12 epochs as the standard training duration.

\begin{table}[h]
    \centering
    \caption{\textbf{Effect of Training Duration.} One epoch corresponds to sampling 100\% of positive pairs. The model saturates at epoch 12.}
    \label{tab:ablation_epoch}
    \begin{tabular}{lc}
        \toprule
        \textbf{Epochs} & \textbf{Test AUC} \\
        \midrule
        10 & 0.8747 \\
        $\mathbf{12}$ & $\mathbf{0.8926}$ \\
        15 & 0.8920 \\
        \bottomrule
    \end{tabular}
\end{table}

\subsection{Impact of Positive-to-Negative Ratio}
Finally, we examine the effect of the dynamic negative sampling ratio ($1:m$). While increasing the number of negative samples typically aids in mining hard negatives, it also alters the class balance and increases training time per epoch. As presented in Table~\ref{tab:ablation_ratio}, the balanced ratio of $1:1$ yields the highest AUC of $0.8926$. Interestingly, increasing the ratio to $1:2$ results in a performance decrease to $0.8837$. This suggests that for our pairwise objective, a balanced contrast between edit-related and unrelated pairs provides the most stable gradient for learning forensic artifacts. Based on this, we adopt a $1:1$ ratio to maintain both high performance and training efficiency.

\begin{table}[h]
    \centering
    \caption{\textbf{Effect of Positive-to-Negative Ratio.} Comparing balanced sampling ($1:1$) versus increased negative sampling ($1:2$). Balanced sampling yields superior results.}
    \label{tab:ablation_ratio}
    \begin{tabular}{lc}
        \toprule
        \textbf{Ratio ($Pos:Neg$)} & \textbf{Test AUC} \\
        \midrule
        $\mathbf{1:1}$ & $\mathbf{0.8926}$ \\
        $1:2$ & 0.8837 \\
        \bottomrule
    \end{tabular}
\end{table}

\section{Dataset Statistics and Distribution Details}
\label{app:stats_details}

The \textbf{EditGroup-45k} dataset is constructed to capture diverse and realistic image manipulation behaviors. In total, the dataset contains \textbf{45,546} edit groups, comprising \textbf{52,074} unique real images and \textbf{232,520} manipulated images. In this section, we provide a detailed analysis of the dataset distribution from four perspectives: editing intensity, source composition, group scale, and similarity metrics.

\subsection{Editing Intensity}
To understand the frequency of source image usage, we analyze the number of fake images generated from each unique real image.
\begin{itemize}
    \item \textbf{Single-use (41.51\%):} A large portion of real images (21,615) are used only once.
    \item \textbf{Repeated-use (58.49\%):} The majority of real images serve as sources for multiple edits, with 31.78\% being edited 2--3 times.
    \item \textbf{Heavy-use ($>$30):} A long-tail distribution exists where 2.81\% of real images are used for more than 30 edits.
\end{itemize}

\begin{figure}[h]
    \centering
    \includegraphics[width=0.6\linewidth]{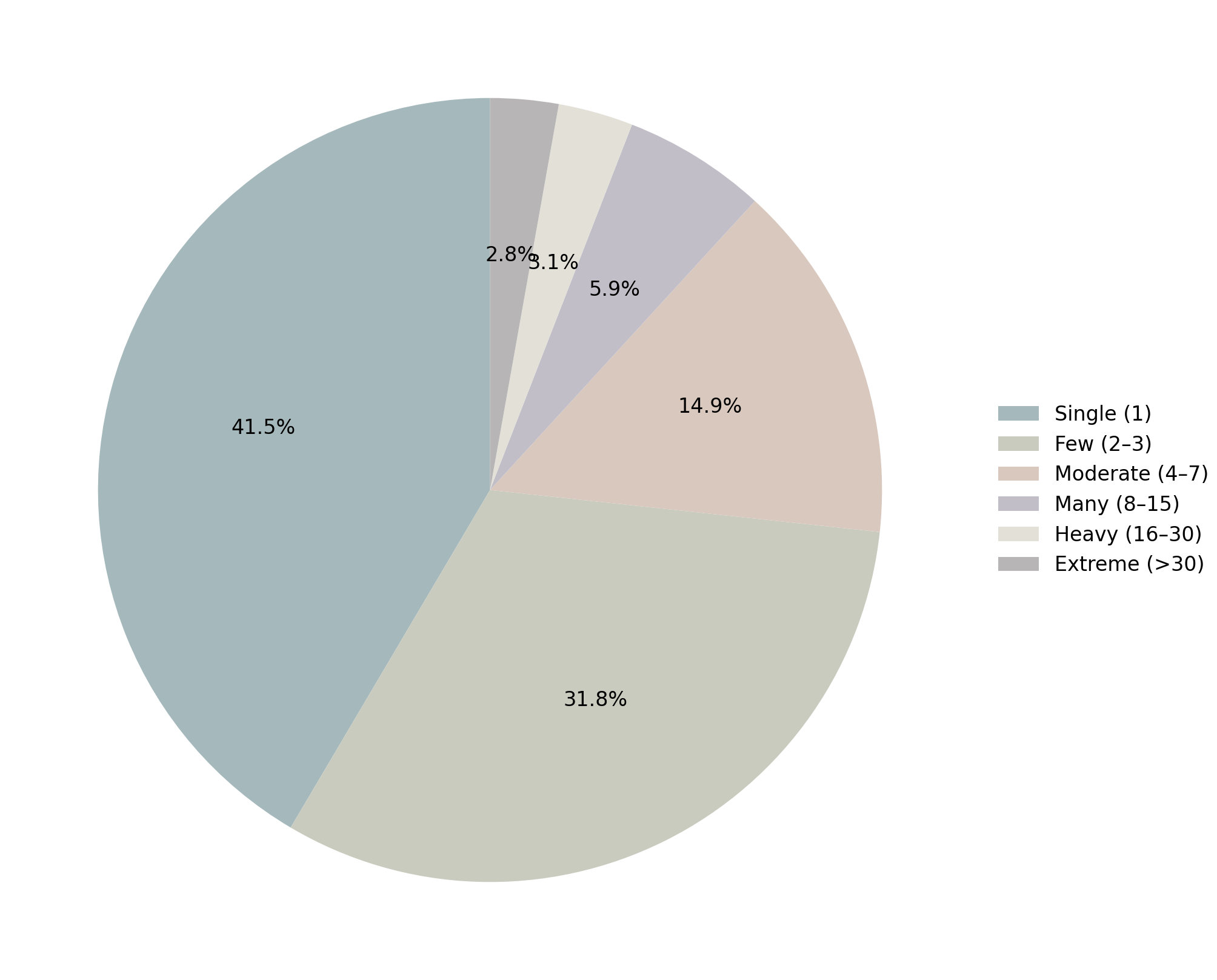}
    \caption{Distribution of edit frequency per real image.}
    \label{fig:real_intensity}
\end{figure}

\subsection{Composition Complexity}
We analyze the composition of manipulated images to determine whether they originate from a single source or multiple sources (splicing).
\begin{itemize}
    \item \textbf{Single-Source (88.69\%):} The vast majority of fake images are derived from a single real image.
    \item \textbf{Multi-Source (11.31\%):} A significant subset involves two or more source images, indicating complex splicing operations.
\end{itemize}

\begin{figure}[h]
    \centering
    \includegraphics[width=0.6\linewidth]{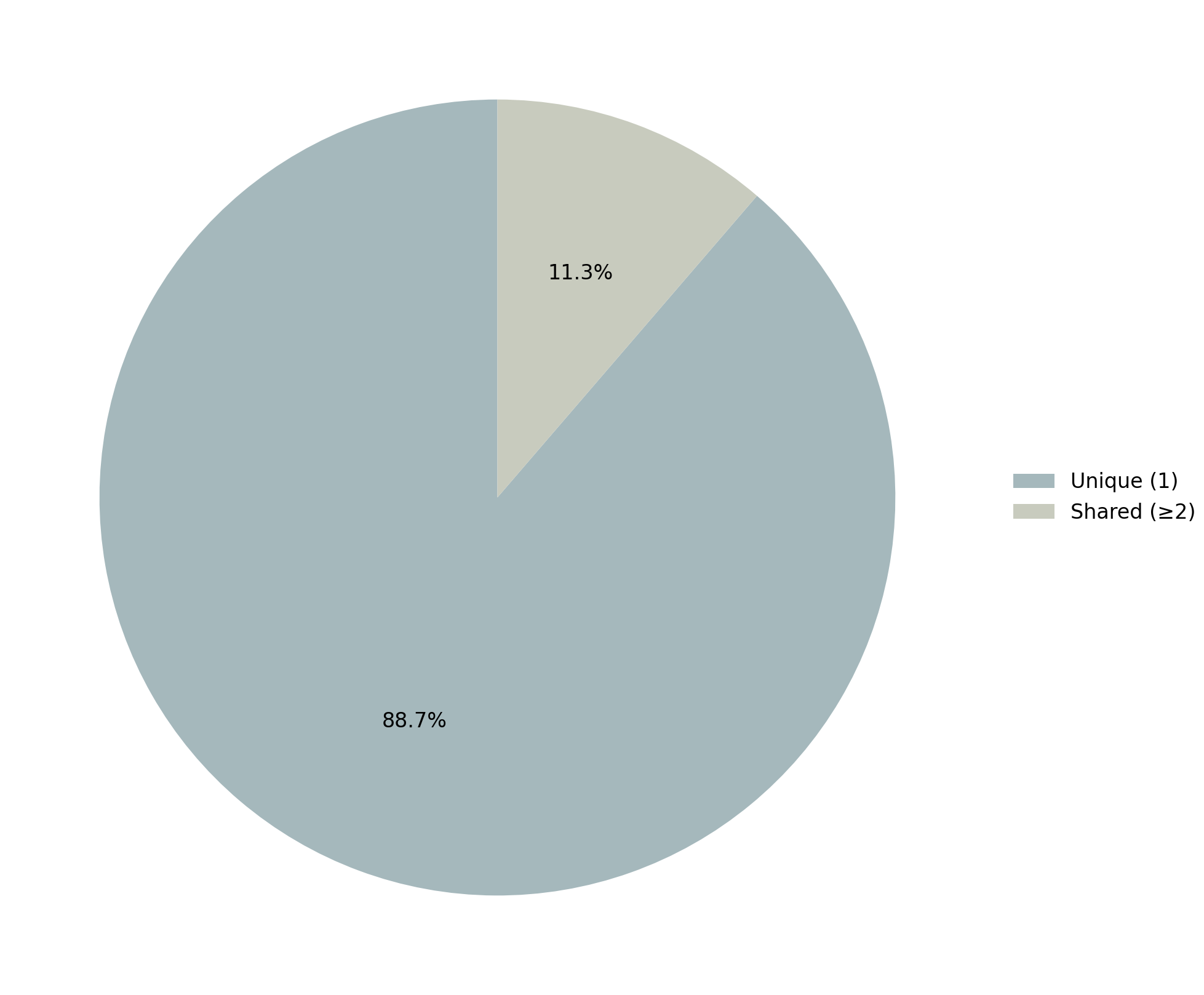}
    \caption{Proportion of single-source vs. multi-source manipulations.}
    \label{fig:fake_reuse}
\end{figure}

\subsection{Group Scale Distribution}
The dataset is organized into ``Edit Groups,'' representing coherent editing sessions.
\begin{itemize}
    \item \textbf{Small Groups (72.76\%):} Most groups contain 1--3 image pairs.
    \item \textbf{Medium to Large Groups:} 18.55\% contain 4--10 pairs, and 5.87\% contain 11--30 pairs.
    \item  \textbf{Extreme(2.8\%):} Few groups contain $>$ 30 image pairs.
\end{itemize}

\begin{figure}[h]
    \centering
    \includegraphics[width=0.6\linewidth]{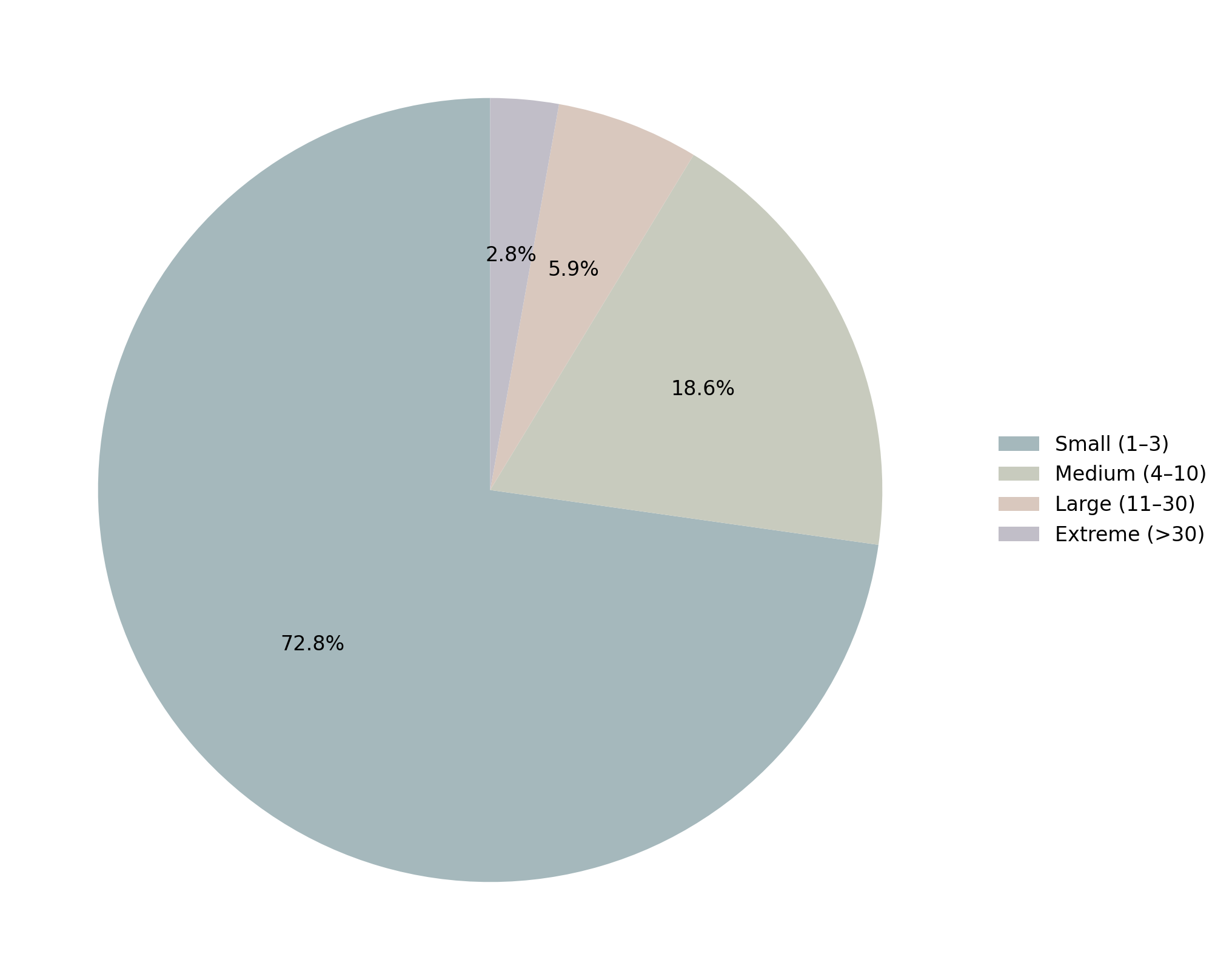}
    \caption{Distribution of image pairs per Edit Group.}
    \label{fig:group_scale}
\end{figure}

\subsection{Similarity and Metric Analysis}
We evaluate the structural and statistical similarity between real and fake pairs to quantify the magnitude of manipulations.

\paragraph{Structural Similarity Levels.}
Based on SSIM values, we categorize the samples into three levels of similarity:
\begin{itemize}
    \item \textbf{High Similarity (SSIM $\ge$ 0.95):} 23.50\% of pairs exhibit high structural preservation.
    \item \textbf{Medium Similarity (0.80--0.95):} 38.11\% represent standard editing ranges.
    \item \textbf{Low Similarity ($<$ 0.80):} 38.40\% involve significant structural changes.
\end{itemize}

\begin{figure}[h]
    \centering
    \includegraphics[width=0.6\linewidth]{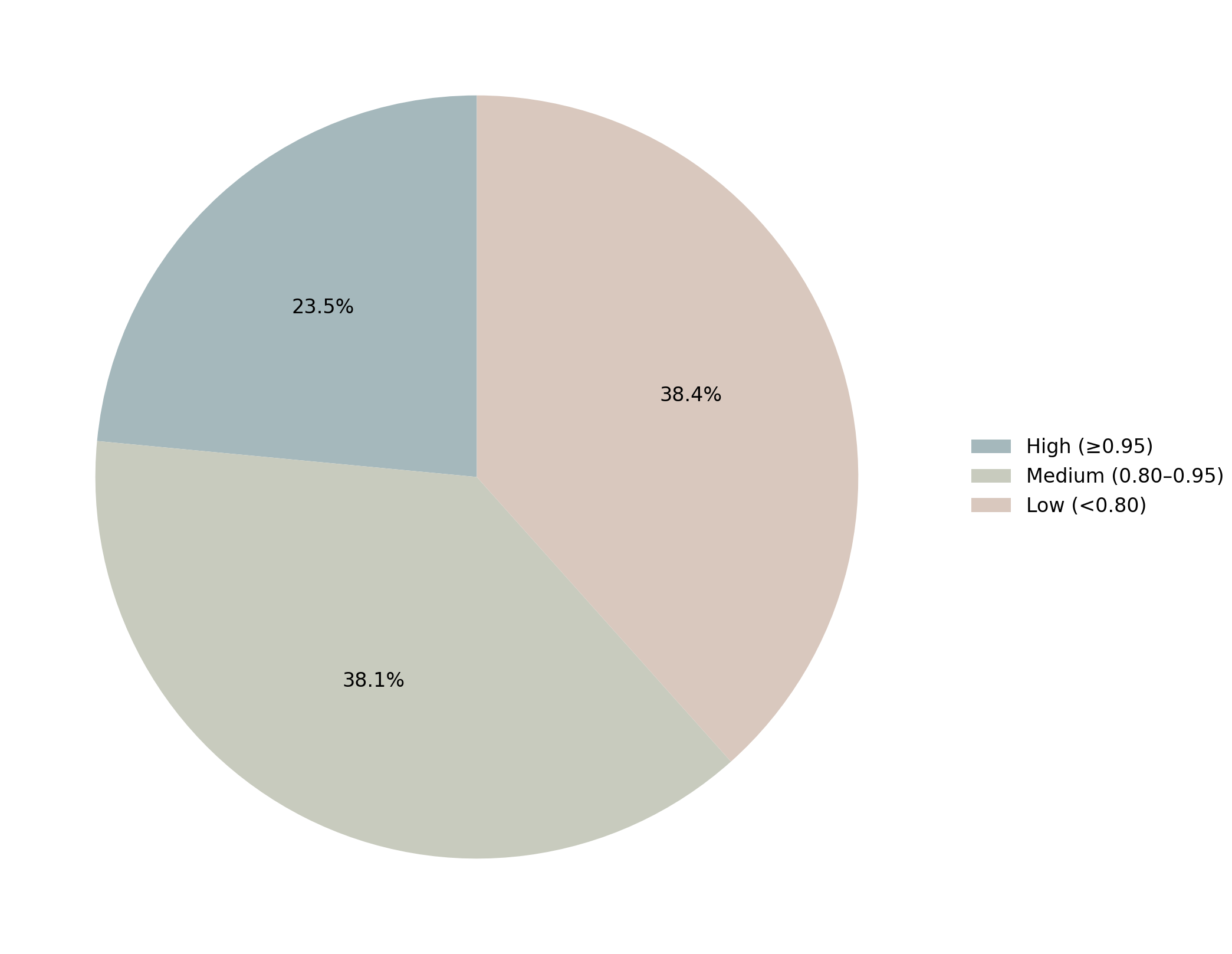}
    \caption{Categorical distribution of Structural Similarity (SSIM).}
    \label{fig:ssim_levels}
\end{figure}

\begin{figure}[h]
    \centering
    \includegraphics[width=0.6\linewidth]{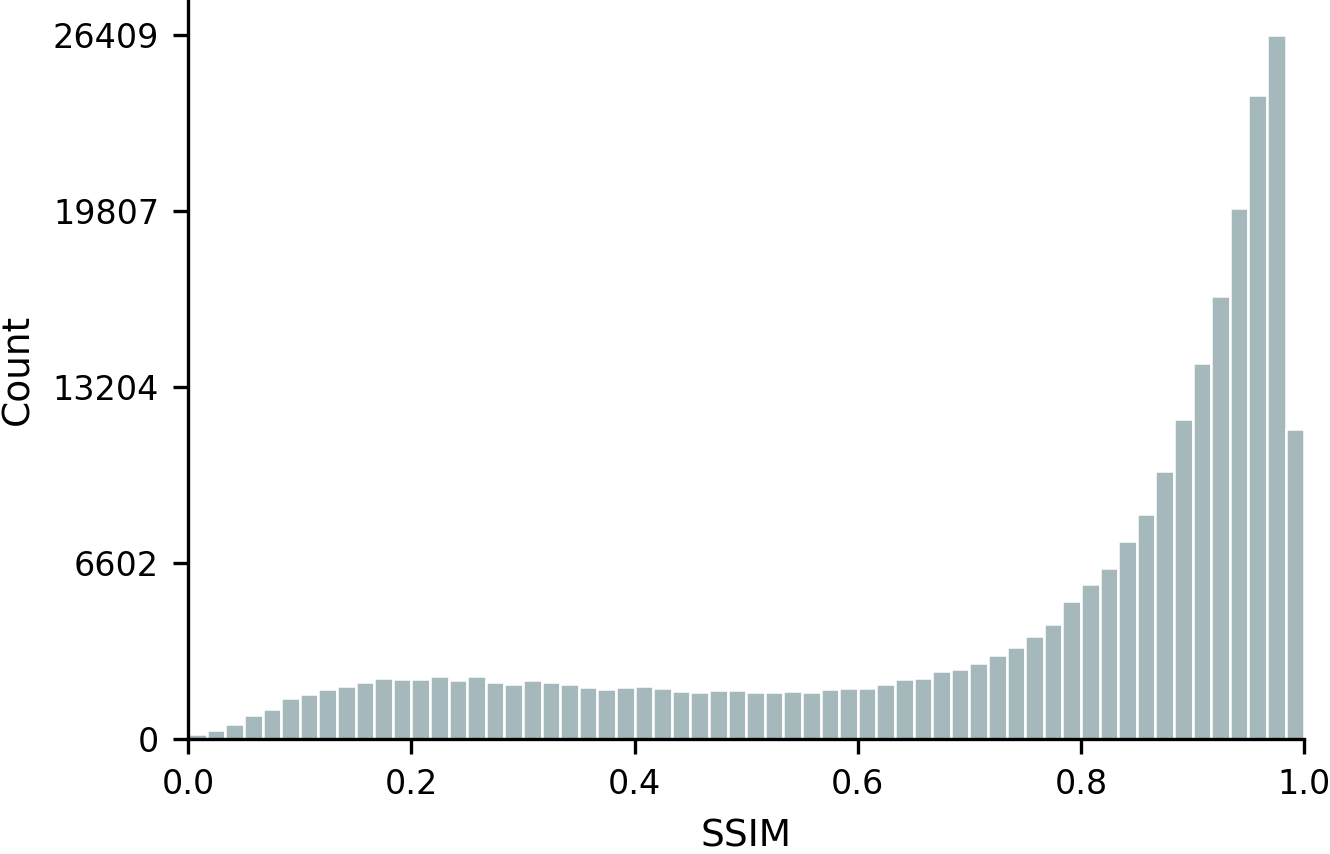} 
    \caption{Continuous probability distribution of SSIM across the dataset.}
    \label{fig:distribution_metrics}
\end{figure}

\paragraph{Continuous Metric Distribution.}
We analyze the continuous probability distributions of similarity and quality metrics to characterize the magnitude of image manipulations, as illustrated in Figure~\ref{fig:distribution_metrics}. The SSIM exhibits a broad variance, covering the spectrum from subtle retouching to significant structural alterations. This statistical diversity confirms that the dataset successfully encapsulates a wide range of realistic editing scenarios, imperceptible modifications with complex structural changes.

\begin{table*}[]
\centering
\caption{Detailed numerical results of the ablation study evaluating different supervision forms, representation designs, and negative pair compositions.
We report the pixel-level F1 score on the Standard Localization task under the Protocol-CAT benchmark.
\textbf{In-Domain} performance is evaluated on the source dataset (CASIAv1), while \textbf{Cross-Domain} performance is measured across five unseen datasets (COVERAGE, Columbia, NIST16, AutoSplice, and COCO-Glide).
The \textbf{Average} column denotes the mean score across the five cross-domain datasets, and the final column (\textbf{All-Avg}) represents the overall mean across all six evaluation datasets.}
\label{tab:full_ablation_appendix}
\resizebox{\linewidth}{!}{
\begin{tabular}{@{}lllllllll@{}}
\toprule
\multicolumn{1}{c}{\multirow{2}{*}{Model}} &
  \multicolumn{1}{c}{In-Domain} &
  \multicolumn{6}{c}{Cross-Domain} &
  \multicolumn{1}{c}{\multirow{2}{*}{Average}} \\ \cmidrule(lr){2-2}\cmidrule(lr){3-8}
\multicolumn{1}{c}{} &
  CASIAv1 &
  COVERAGE &
  Columbia &
  NIST16 &
  Autosplice &
  CocoGlide &
  Average &
  \multicolumn{1}{c}{} \\ \cmidrule(r){1-9} 
Baseline     & 0.7369 & 0.5241 & 0.8976 & 0.3754 & 0.4430  & 0.4198 & 0.5320 & 0.5661 \\
Baseline-ABS & 0.8457 & 0.5888 & 0.9361  & 0.3650  & 0.3637 & 0.3927 & 0.5292 & 0.5820 \\ \midrule
PAL(Ours)    & 0.8256 & 0.6242 & 0.9043 & 0.3849 & 0.4830 & 0.5375 & 0.5868 & 0.6266 \\ \midrule
w/o Diff     & 0.8073 & 0.426  & 0.9233 & 0.3763 & 0.4794 & 0.4617 & 0.5333 & 0.5790 \\
w/o AvgPool  & 0.7814 & 0.5826 & 0.8618 & 0.3312 & 0.4415 & 0.5017 & 0.5438 & 0.5834 \\
w/o MaxPool  & 0.8368 & 0.6084 & 0.9136 & 0.3462 & 0.4579 & 0.5118 & 0.5676 & 0.6125 \\ \midrule
w/o CG-OE    & 0.8232 & 0.5459 & 0.9165 & 0.3439 & 0.4825 & 0.4729 & 0.5523 & 0.5975 \\
w/o SA       & 0.8521 & 0.5640 & 0.9184 & 0.3778 & 0.4169 & 0.4691 & 0.5492 & 0.5997 \\
w/o IG-EE    & 0.7893 & 0.4258 & 0.9271 & 0.2992 & 0.5273 & 0.5137 & 0.5386 & 0.5804 \\
w/o IG-OO    & 0.8239 & 0.5454 & 0.8662 & 0.3553 & 0.4738 & 0.5287 & 0.5539 & 0.5989 \\
w/o  G-EE    & 0.7790 & 0.5403 & 0.9460 & 0.3574 & 0.4694 & 0.4426 & 0.5511 & 0.5891 \\
w/o  G-OO    & 0.8183 & 0.5650 & 0.9238 & 0.3692 & 0.4887 & 0.5031 & 0.5700 & 0.6114 \\ \bottomrule
\end{tabular}
}
\end{table*}

\section{Supplementary Ablation Study}
\label{app:ablation}

In this supplementary section, we provide a comprehensive ablation study to further validate the effectiveness of the PAL framework and its key design components.
While the main paper discusses the aggregate trends, here we present detailed performance metrics across all individual datasets to offer deeper insights into the behavior of each variant.
It is important to note that all reported metrics reflect the final performance on the \textbf{Standard Localization} task under the Protocol-CAT benchmark.
To ensure fair comparison, all variants utilize the same ConvNeXt-Small backbone (pretrained on ImageNet-1K) and undergo identical Standard Localization training; differences among variants originate solely from the supervision strategies or design choices within the PAL stage.
The complete numerical results are summarized in Table~\ref{tab:full_ablation_appendix}.

\subsection{Effect of PAL}
\label{app:ablation_supervision}

We first compare different supervision forms to analyze the trade-off between source domain fitting and cross-domain generalization.
\textbf{Baseline} is initialized with ImageNet weights only.
\textbf{Baseline-ABS} applies absolute binary supervision on the EditGroup-45K dataset without pairwise relations.
\textbf{PAL} employs our proposed pairwise learning strategy to disentangle manipulation artifacts.

As shown in Table~\ref{tab:full_ablation_appendix}, \textbf{PAL} achieves the highest All-Avg of 0.6266, demonstrating superior overall robustness.
A granular analysis reveals the limitations of absolute supervision.
While \textbf{Baseline-ABS} achieves remarkable performance on the source domain CASIAv1 (0.8457) and the relatively simpler Columbia dataset (0.9360), its performance deteriorates significantly on more complex benchmarks.
Specifically, on the challenging NIST16 and AutoSplice datasets, Baseline-ABS yields scores of 0.3650 and 0.3637 respectively, which are inferior even to the vanilla Baseline (0.3754 and 0.4430).
This trend indicates that absolute supervision tends to overfit to the source distribution, failing to capture transferable forensic features.
In contrast, \textbf{PAL} demonstrates superior robustness, effectively recovering performance on these hard datasets—improving AutoSplice to 0.4830 and CocoGlide to 0.5375—thereby validating that pairwise modeling learns generic representations resilient to unseen manipulation patterns.

\subsection{Effect of Pair Representation Design}
\label{app:ablation_representation}

We next evaluate the design of the pair representation module within the PAL framework.
\textbf{w/o Diff} removes the explicit feature difference stream, while \textbf{w/o AvgPool} and \textbf{w/o MaxPool} exclude global average and max pooling, respectively.

The removal of explicit difference modeling (\textbf{w/o Diff}) results in the most significant degradation in generalization, with the Cross-Avg dropping from 0.5868 to 0.5333.
This decline is particularly pronounced on datasets requiring fine-grained trace detection, such as COVERAGE, where the score plummets to 0.4260 compared to 0.6242 for the full model.
This confirms that the explicit difference map is the cornerstone for extracting generalized manipulation artifacts.
Regarding pooling strategies, the exclusion of average pooling (\textbf{w/o AvgPool}) leads to the lowest In-Domain performance on CASIAv1 (0.7814), suggesting that global context aggregation is essential for stability in the source domain.
Conversely, while \textbf{w/o MaxPool} maintains competitive in-domain scores, its generalization capability is weakened across cross-domain benchmarks, indicating that capturing peak local inconsistencies is vital for cross-domain robustness.

\begin{table*}[]
\centering
\caption{Robustness analysis under common post-processing perturbations.
We evaluate the robustness of the Baseline and Baseline+PAL models under Gaussian blur with increasing kernel sizes and JPEG compression with decreasing quality factors.
PAL consistently improves robustness across all perturbation levels, exhibiting significantly slower performance degradation as perturbations intensify.
The reported F1 score (rightmost column) summarizes the average performance over each perturbation setting.}
\label{tab:robustness}
\resizebox{\linewidth}{!}{
\begin{tabular}{@{}clllllllll@{}}
\toprule
\multirow{2}{*}{Perturbation} & \multicolumn{1}{c}{\multirow{2}{*}{Model}} & \multicolumn{7}{c}{Kernel Size}    & \multicolumn{1}{c}{\multirow{2}{*}{F1}} \\ \cmidrule(lr){3-9}
                                 & \multicolumn{1}{c}{} & None   & 3      & 7      & 11     & 15     & 19     & 23     & \multicolumn{1}{c}{} \\ \midrule
\multirow{2}{*}{GaussBlur}       & Baseline             & 0.7369 & 0.6690 & 0.5853 & 0.4938 & 0.3741 & 0.1746 & 0.0490 & 0.4404               \\
                                 & Baseline+PAL         & 0.8256 & 0.7590 & 0.7166 & 0.6324 & 0.5157 & 0.3619 & 0.2030 & 0.5735               \\ \midrule
\multirow{2}{*}{}            & \multicolumn{1}{c}{\multirow{2}{*}{}}      & \multicolumn{7}{c}{QualityFactors} & \multicolumn{1}{c}{\multirow{2}{*}{}}   \\ \cmidrule(lr){3-9}
                                 & \multicolumn{1}{c}{} & None   & 100    & 90     & 80     & 70     & 60     & 50     & \multicolumn{1}{c}{} \\ \midrule
\multirow{2}{*}{JpegCompression} & Baseline             & 0.7369 & 0.7124 & 0.6880 & 0.6301 & 0.6268 & 0.5943 & 0.5038 & 0.6418               \\
                                 & Baseline+PAL         & 0.8256 & 0.8128 & 0.7992 & 0.7466 & 0.7298 & 0.6901 & 0.6127 & 0.7452               \\ \bottomrule
\end{tabular}
}
\end{table*}
\subsection{Effect of Negative Pair Composition}
\label{app:ablation_negative}

Finally, we analyze the impact of specific negative pair types, denoting Original as 'O' and Edited as 'E'. The variants include removing Cross-group Original--Edited (\textbf{w/o CG-OE}), In-group pairs (\textbf{w/o IG-OO/EE}), Global pairs (\textbf{w/o G-OO/EE}), and Self-Augmented negatives (\textbf{w/o SA}).

The removal of In-group Edited--Edited pairs (\textbf{w/o IG-EE}) proves most detrimental to generalization, resulting in the lowest Cross-Avg of 0.5386.
This impact is starkly visible on the NIST16 dataset, where performance collapses to 0.2992, confirming that distinguishing between different manipulation types within the same semantic context is vital for recognizing diverse artifacts in unseen domains.
Conversely, removing Global Edited--Edited pairs (\textbf{w/o G-EE}) primarily harms source domain coverage, yielding the lowest CASIAv1 score of 0.7790.

Most notably, removing Self-Augmented negatives (\textbf{w/o SA}) causes the In-Domain score on CASIAv1 to spike to 0.8521, surpassing the full PAL model.
However, this apparent gain is accompanied by a severe degradation in cross-domain performance, with scores on AutoSplice dropping to 0.4169 compared to 0.4830 for PAL.
This sharp discrepancy indicates that without the constraint of self-augmentation, the model learns trivial shortcuts specific to the source data, leading to severe overfitting and a loss of robustness against unseen manipulations.

\subsection{Robustness to Post-processing Perturbations.}
To evaluate robustness against common post-processing operations that tend to obscure low-level forensic cues, we assess model performance on \textbf{CASIA v1}~\cite{CASIA_2013} under Gaussian blur and JPEG compression~\cite{ma2023iml}, as summarized in Table~\ref{tab:robustness}.
Both perturbations progressively suppress manipulation traces, with increasing blur kernel sizes or decreasing JPEG quality factors leading to a monotonic performance degradation for all models.

Across all perturbation levels, \textbf{Baseline+PAL} consistently outperforms the Baseline by a large margin.
Under Gaussian blur, PAL exhibits substantially slower degradation as kernel size increases, maintaining usable performance even under severe blurring, where the Baseline nearly collapses.
A similar trend is observed for JPEG compression, where PAL preserves higher F1 scores across all quality factors, particularly under aggressive compression.

These results indicate that PAL learns more stable and transferable artifact representations that are less sensitive to post-processing distortions.
By explicitly modeling relational discrepancies rather than relying on fragile absolute cues, PAL improves robustness to operations that partially destroy or smooth manipulation artifacts.

\begin{figure}[h]
    \centering
    \includegraphics[width=0.6\linewidth]{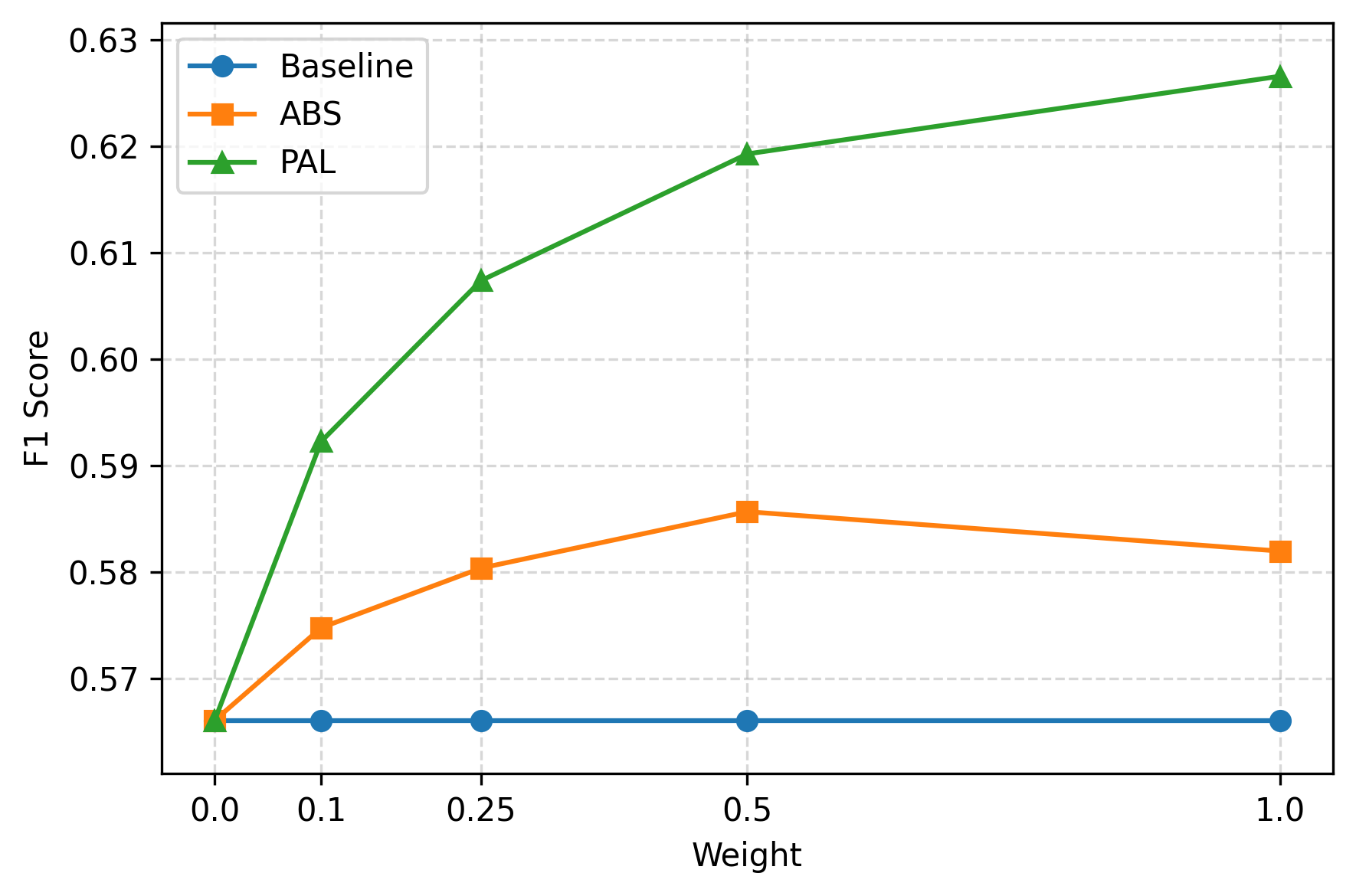} 
    \caption{Scaling behavior with different pre-training ratios of EditGroup-45K.
    PAL benefits consistently from a larger data scale, while ABS saturates and slightly degrades when using the full dataset.}
    \label{fig:scaling_behavior}
\end{figure}

\subsection{Scaling Behavior of EditGroup-45K}
\label{app:scaling_behavior}

To further examine whether the improvement of PAL comes from the dataset scale or the pairwise learning objective, we evaluate different pre-training ratios of EditGroup-45K, including 10\%, 25\%, 50\%, and 100\%.
As shown in Fig.~\ref{fig:scaling_behavior}, PAL shows consistent performance gains as the amount of pre-training data increases, indicating that pairwise relation learning can effectively benefit from larger-scale edit groups.
In contrast, absolute supervision (ABS) improves at the early stage but reaches a performance ceiling around the 50\% subset and slightly degrades when using the full dataset.
This suggests that directly assigning binary labels to individual images is less compatible with the source-anchored setting, where some anchor images may already contain prior editing or generation traces before the target manipulation.
Such cases can introduce label conflicts for ABS, since these images are forced to be treated as pristine sources.
PAL avoids this issue by learning relative edit differences within pairs, making it more scalable under heterogeneous source conditions.

\begin{figure}[t]
    \centering
    \includegraphics[width=0.6\linewidth]{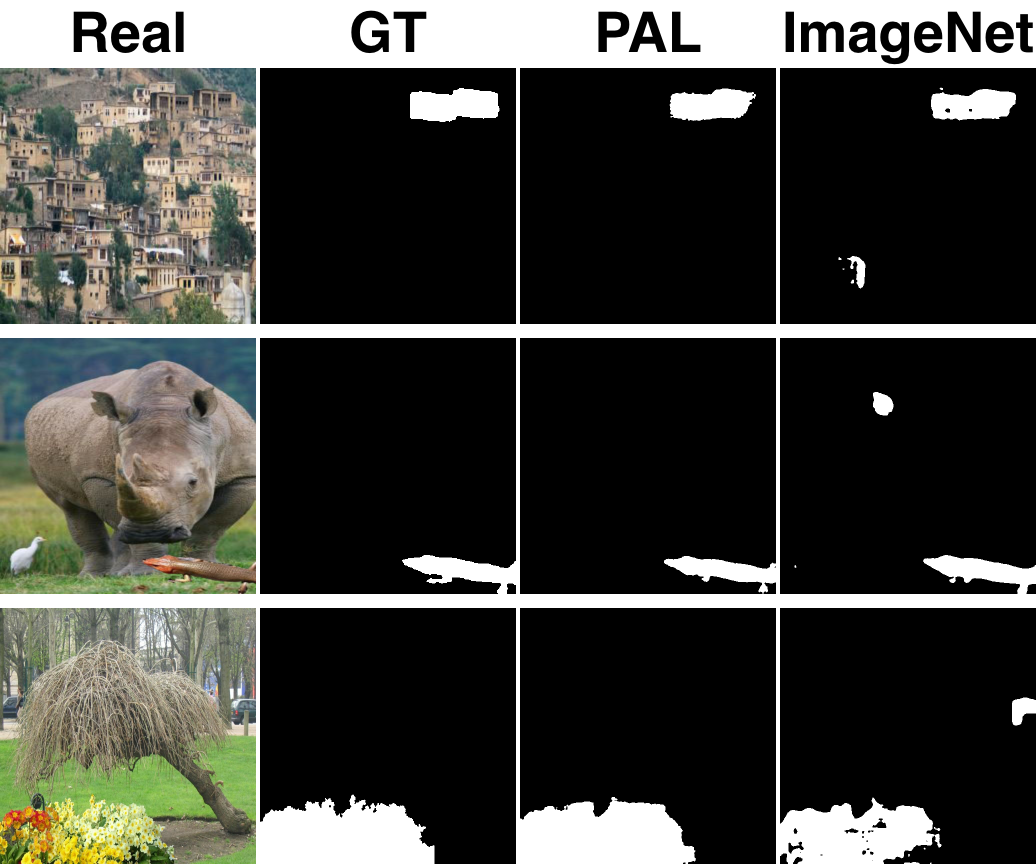}
    \caption{\textbf{Qualitative comparison between PAL and ImageNet initialization.}
    Rows show representative copy-move, splicing, and inpainting examples.
    PAL produces localization results closer to GT and fewer false positives than ImageNet initialization.}
    \label{fig:pal_imagenet_vis}
\end{figure}

\subsection{Qualitative Comparison between PAL and ImageNet Initialization}
\label{app:pal_imagenet_vis}

We further compare the localization results obtained with PAL initialization and ImageNet initialization.
As shown in Fig.~\ref{fig:pal_imagenet_vis}, the three rows correspond to representative manipulation types, including copy-move, splicing, and inpainting.
Compared with ImageNet initialization, PAL produces masks that are more consistent with the ground truth and introduces fewer false positive regions.
This indicates that PAL provides an artifact-aware initialization that helps the downstream localization model focus on manipulation-related regions rather than semantic or background distractors.

\section{Broader Impact}
\label{app:broader_impact}
The scope of PAL is not limited to manipulation localization in 2D natural images. At its core, PAL seeks to explicitly model manipulation artifacts as latent representations and disentangle them from semantic content, rather than directly learning the mapping from images to manipulation labels. Such an artifact-oriented formulation is not inherently tied to RGB images, and may therefore provide a general basis for extending forensic representation learning to more diverse content modalities.

One promising direction is to generalize PAL toward visual domains with richer structural representations. For remote sensing imagery, recent advances have demonstrated the importance of multistream spatial modeling~\cite{10945380} and spectral--spatial representation learning~\cite{11418730}, suggesting that manipulation artifacts could be explored beyond conventional RGB traces toward spectral and spatial inconsistencies. A similar extension can be envisioned for digital humans, where language--pose pretraining~\cite{Jia_2026_CVPR}, in-the-wild motion recovery~\cite{jia2026ramrecover3dhuman}, and multimodal motion--language alignment~\cite{jia2026unified} provide structured representations of human dynamics. Recent studies on motion instruction tuning~\cite{li2025human} and multi-person motion understanding~\cite{li2026multiple} further indicate that human motion can be represented at increasingly fine-grained semantic and interaction levels. Extending PAL to such representations may enable manipulation artifacts to be modeled not only as local appearance anomalies, but also as inconsistencies in pose, motion, and human interactions. Beyond human-centric content, geometric reconstruction methods such as Niagara~\cite{11299106} also provide a potential foundation for investigating artifact representations associated with geometric deformation and structural inconsistency in 3D content.

Another direction is to enrich PAL with more adaptive mechanisms for organizing and exploiting learned artifact representations. Query-aware evidence retrieval~\cite{jiang2025qa} and vision-language agentic perception~\cite{jiang2026superglasses} suggest the potential of selecting complementary evidence according to the input content, while hierarchical structure reasoning~\cite{jiang2025hibench} provides a possible means of organizing forensic evidence at multiple levels of abstraction. More broadly, recent developments in cross-domain semantic reasoning~\cite{jiang2026atomic} and self-evolving strategy optimization~\cite{jiang2026b} motivate exploring whether the downstream use of PAL representations can be dynamically adapted to different manipulation types and domains.

Finally, PAL may also be extended from manipulation forensics toward broader trustworthy-content protection. Recent studies on transferable reversible adversarial examples for image privacy protection~\cite{du2025dp} and semantics-driven defensive adversarial CAPTCHA generation~\cite{du2025defensive} demonstrate the potential of explicitly modeling robust visual perturbations for security-oriented tasks. Meanwhile, embedding-space watermarking under semantic-invariant attacks~\cite{ai2026pasa} further highlights the need for representations that remain informative under substantial content-preserving transformations. These directions suggest that artifact-oriented representation learning could support not only manipulation detection, but also privacy protection, adversarial defense, and content authentication. Overall, extending PAL toward spectral, motion, geometric, and security-oriented representations could help establish a more general paradigm for transferable forensic representation learning.

\section{Limitations}
\label{app:limitations}

This work focuses on a general artifact-aware initialization paradigm rather than proposing a specific IML model architecture.
Therefore, PAL does not customize the backbone, decoder, or localization head for any particular model, which may limit model-specific optimization.
Future work can further explore dedicated architectures that integrate pairwise artifact learning more deeply into the localization process.






  \newpage
\end{document}